\pdfoutput=1

\documentclass[11pt]{article}

\usepackage{EACL2023}

\usepackage{times}
\usepackage{latexsym}

\usepackage[T1]{fontenc}

\usepackage[utf8]{inputenc}

\usepackage{microtype}

\usepackage{inconsolata}

\usepackage{xcolor}
\usepackage{multirow}
\usepackage{array,booktabs}
\usepackage{bm}
\usepackage{amssymb}
\usepackage{amsmath}
\usepackage{cleveref}
\usepackage{mathtools}
\usepackage{xspace}
\usepackage{tikz}
\usepackage{soul}
\usepackage{subcaption}

\DeclareMathOperator*{\argmax}{arg\,max}

\crefformat{section}{\S#2#1#3} 
\crefformat{subsection}{\S#2#1#3}

\newcommand{\reducedstrut}{\vrule width 0pt height 1.2\ht\strutbox depth 1.2\dp\strutbox\relax}
\newcommand{\SU}[1]{%
  \begingroup
  \setlength{\fboxsep}{0pt}%
  \colorbox{blue!30}{\reducedstrut#1\/}%
  \endgroup
}
\newcommand{\SUcaption}[1]{%
  \begingroup
  \setlength{\fboxsep}{0pt}%
  \colorbox{blue!30}{\strut#1\/}%
  \endgroup
}

\newcommand{\roberta}{RoBERTa$_{\small \textsc{BASE}}$\xspace}

\makeatletter
\renewcommand{\maketag@@@}[1]{\hbox{\m@th\normalsize\normalfont#1}}%
\makeatother

\title{Sentence Identification with BOS and EOS Label Combinations}

\author{Takuma Udagawa, Hiroshi Kanayama, Issei Yoshida \\
        IBM Research - Tokyo, Japan \\
        \texttt{Takuma.Udagawa@ibm.com, \{hkana,issei\}@jp.ibm.com}}

\begin{document}
\maketitle
\begin{abstract}
The sentence is a fundamental unit in many NLP applications.
Sentence segmentation is widely used as the first preprocessing task, where an input text is split into consecutive sentences considering the end of the sentence (EOS) as their boundaries.
This task formulation relies on a strong assumption that the input text consists only of sentences, or what we call the sentential units (SUs).
However, real-world texts often contain non-sentential units (NSUs) such as metadata, sentence fragments, nonlinguistic markers, etc. which are unreasonable or undesirable to be treated as a part of an SU.
To tackle this issue, we formulate a novel task of sentence identification, where the goal is to identify SUs while excluding NSUs in a given text.
To conduct sentence identification, we propose a simple yet effective method which combines the beginning of the sentence (BOS) and EOS labels to determine the most probable SUs and NSUs based on dynamic programming.
To evaluate this task, we design an automatic, language-independent procedure to convert the Universal Dependencies corpora into sentence identification benchmarks.
Finally, our experiments on the sentence identification task demonstrate that our proposed method generally outperforms sentence segmentation baselines which only utilize EOS labels.

\end{abstract}

\section{Introduction}

\begin{table*}[t]
\centering \scalebox{0.81}{
\setlength{\aboverulesep}{1pt}
\setlength{\belowrulesep}{1pt}
\begin{tabular}{cl}
\toprule[\heavyrulewidth]
Input Text & \texttt{Thank you. - TEXT.htm {\textless}{\textless} File: TEXT.htm {\textgreater}{\textgreater} I was thinking of converting it to a hover} \\
(from EWT) & \texttt{vehicle. I might just sell the car and get you to drive me around all winter.} \\
\midrule
\midrule
 & \phantom{\texttt{Thank you}}\texttt{\textbf{E}}\phantom{\texttt{ - TEXT.htm {\textless}{\textless} File: TEXT.htm {\textgreater}{\textgreater} I was thinking of converting it to a hover}} \\
Sentence & \texttt{\SU{Thank you.} \SU{- TEXT.htm {\textless}{\textless} File: TEXT.htm {\textgreater}{\textgreater} I was thinking of converting it to a hover}} \\
Segmentation& \phantom{\texttt{vehicle}}\texttt{\textbf{E}}\phantom{\texttt{ I might just sell the car and get you to drive me around all winter}}\texttt{\textbf{E}} \\
& \texttt{\SU{vehicle.} \SU{I might just sell the car and get you to drive me around all winter.}} \\
\midrule
 & \texttt{\textbf{B}}\phantom{\texttt{hank you}}\texttt{\textbf{E}}\phantom{\texttt{ - TEXT.htm {\textless}{\textless} File: TEXT.htm {\textgreater}{\textgreater} }}\texttt{\textbf{B}}\phantom{\texttt{ was thinking of converting it to a hover}} \\
Sentence & \texttt{\SU{Thank you.} - TEXT.htm {\textless}{\textless} File: TEXT.htm {\textgreater}{\textgreater} \SU{I was thinking of converting it to a hover}} \\
Identification& \phantom{\texttt{vehicle}}\texttt{\textbf{E} }\texttt{\textbf{B}}\phantom{\texttt{ might just sell the car and get you to drive me around all winter}}\texttt{\textbf{E}} \\
& \texttt{\SU{vehicle.} \SU{I might just sell the car and get you to drive me around all winter.}} \\
\bottomrule[\heavyrulewidth]
\end{tabular}
}
\caption{Illustration of sentence segmentation and sentence identification. In sentence segmentation, EOS labels (\texttt{\textbf{E}}) are used to segment the input text into consecutive SUs (in \SUcaption{blue}). In sentence identification, only the spans bracketed by the BOS (\texttt{\textbf{B}}) and EOS labels are extracted as SUs, while the rest can be excluded as NSUs.
}
\label{table:si_illustration}
\end{table*}

The sentence, which we refer to as the sentential unit (SU), is a fundamental unit of processing in many NLP applications including syntactic parsing \citep{dozat2017deep}, semantic parsing \citep{dozat-manning-2018-simpler}, and machine translation \citep{liu-etal-2020-multilingual-denoising}. 
Existing works mostly rely on \textit{sentence segmentation} (a.k.a. \textit{sentence boundary detection}) as the first preprocessing task, where we predict the end of the sentence (EOS) to split a text into consecutive SUs \citep{kiss-strunk-2006-unsupervised,gillick-2009-sentence}.
This approach relies on a strong assumption that the text only consists of SUs; however, real-world texts like web contents often contain non-sentential units (NSUs) such as the metadata of attachments embedded in the email body, repetition of symbols for separating texts, irregular series of nouns, etc. (just to name a few).
Such NSUs may cause detrimental or unexpected results in the downstream tasks if considered as parts of the SUs and are more desirable to be distinguished from SUs in the first preprocessing step.

To tackle this problem, we formulate a novel task of \textit{sentence identification}, where the goal is to identify SUs while excluding NSUs in a given text (\cref{sec:task_formulation}).
This can be regarded as an SU span extraction task, where each SU span is represented by the beginning of the sentence (BOS) and the EOS labels.\footnote{For simplicity, we assume that the input text can be segmented into consecutive, non-overlapping units of SUs and NSUs. This way, we can also represent and evaluate SU extraction as an equivalent BIO labeling task (\cref{sec:evaluation}-\cref{sec:experiment_results}).}
We illustrate the difference between sentence segmentation and sentence identification in Table \ref{table:si_illustration}.
In sentence segmentation, the text fragment of an embedded file (``\texttt{- TEXT.htm {\textless}{\textless} File: TEXT.htm {\textgreater}{\textgreater}}'') needs to be considered as a part of an SU.
In contrast, sentence identification can regard it as an NSU and exclude it for downstream applications such as dependency parsing.

To conduct sentence identification, we propose a simple method which effectively combines the BOS and EOS probabilities to determine both SUs and NSUs (\cref{sec:methods}).
To be specific, we first train the BOS and EOS labeling models based on either the sentence identification dataset (with SUs and NSUs) or sentence segmentation dataset (only SUs).
Then, we search for the most probable spans of SUs and NSUs using a simple dynamic programming framework.
Theoretically, our method can be considered as a natural generalization of existing sentence segmentation algorithms.

To evaluate this task, we design an automatic procedure to convert the Universal Dependencies (UD) corpora \citep{de-marneffe-etal-2021-universal} into sentence identification benchmarks (\cref{sec:evaluation}).
To be specific, (i) we use the original sentence boundaries in UD as the unit (SU and NSU) boundaries and (ii) classify each unit as an SU iff it contains at least one clausal predicate with a core/non-core argument.
Importantly, our classification rule follows the definition of \textit{lexical sentence} in linguistics \citep{nunberg1990linguistics}, is easily customizable with language-independent rules, and makes reasonable classification within the scope of our experiments.

To conduct our experiments, we focus on the English Web Treebank \citep{silveira-etal-2014-gold} as the primary benchmark for sentence identification and train the BOS/EOS labeling models by finetuning RoBERTa \citep{liu2019roberta} (\cref{sec:experimental_setup}).
We also propose techniques to develop these models using a standard sentence segmentation dataset, i.e. the Wall Street Journal corpus \citep{marcus-etal-1993-building}, which only contains clean, edited SUs without any NSUs.

Based on our experimental results, we demonstrate that our proposed method generally outperforms sentence segmentation baselines which only utilize EOS labels (\cref{sec:experiment_results}). These results highlight the importance of combining the BOS labels in addition to the EOS labels for accurate sentence identification under various conditions.

\section{Background}
\label{sec:background}

Sentence segmentation, a.k.a. sentence boundary detection, is the task of segmenting an input text into the unit of sentences.
Despite the long history of study \citep{riley1989some} and its importance in the entire NLP pipeline \citep{walker2001sentence}, this area has received relatively little attention.
For one reason, the task has been recognized as ``long solved'' \citep{read-etal-2012-sentence} with the most recent approach reporting 99.8\% F1 score on the standard English Wall Street Journal (WSJ) dataset \citep{wicks-post-2021-unified}.
Their state-of-the-art method \textsc{Ersatz} combines (i) a regular-expression based detector of candidate sentence boundaries, followed by (ii) a Transformer-based \citep{vaswani2017attention} binary classifier which predicts whether the candidate boundary is EOS based on the local context, i.e. surrounding few words.
This modern context-based approach has been shown to outperform competitive, widely used baselines such as \textsc{Splitta} \citep{gillick-2009-sentence}, \textsc{Punkt} \citep{kiss-strunk-2006-unsupervised}, and \textsc{Moses} \citep{koehn-etal-2007-moses}.

However, two important aspects are not fully addressed in the current literature.
First is the coverage of \textit{diverse domains, genres, and writing styles}.
Existing works (including \citealp{wicks-post-2021-unified}) focus on formal/edited text and assume the existence of sentence ending punctuations (e.g. full stops) at the sentence boundaries.
However, social media texts often lack such punctuations and contain various types of non-linguistic noise, which can lead to a substantial degradation in the segmentation performance \citep{read-etal-2012-sentence,rudrapal-etal-2015-sentence}.
Speech transcription texts also usually contain disfluent, ungrammatical, or fragmented structures and lack both punctuations and casing \citep{wang-etal-2019-online,rehbein-etal-2020-improving}.
Considering the amount of such informal or non-standard texts in the real world, it is compelling to expand the capability of sentence segmentation beyond formal, standardized text.

The second aspect is the coverage of \textit{multiple languages}.
Different languages involve different complexities in sentence segmentation, e.g. Chinese requires the disambiguation of commas as the sentence ending punctuation \citep{xue-yang-2011-chinese} and Thai does not mark EOS with any type of punctations \citep{aroonmanakun2007thoughts,zhou-etal-2016-word}.
To advance NLP from a multilingual perspective, it is crucial to develop and evaluate models in multiple languages: \citet{wicks-post-2021-unified} make an important step in this direction, proposing a language-agnostic, unified sentence segmentation model covering a total of 87 languages.

Based on these observations, we first propose to extend the task of sentence segmentation to \textit{sentence identification}, which expands the capability of sentence segmentation beyond formal, standardized text (\cref{sec:task_formulation}, \cref{sec:methods}).
Secondly, we propose a cross-lingual method of benchmarking sentence identification based on the UD corpora, considering every word or character as the candidate boundary to cover diverse domains, genres, and languages that lack sentence ending punctuations (\cref{sec:evaluation}).
Finally, we follow \citet{wicks-post-2021-unified} to develop modern neural-based models that require no language-specific engineering and can be developed for different languages in a unified manner (\cref{sec:experimental_setup}).

\section{Task Formulation}
\label{sec:task_formulation}

\subsection{Sentence Segmentation Task}
\label{subsec:sentence_segmentation_task}

First, we introduce a precise (re-)formulation of the sentence segmentation task.
Let $\bm{W} = (w_0, w_1,..., w_{N-1})$ represent the input text, where each $w_i$ denotes a word (but can also be a subword or character).
We also define the text span $\bm{W}[i\!:\!j] = (w_{i},..., w_{j-1})$, their concatenation $\bm{W}[i\!:\!j] \oplus \bm{W}[j\!:\!k] = \bm{W}[i\!:\!k]$, and SU boundary indices $\bm{B} = (b_0, b_1,..., b_{M})$ where $b_0 = 0$, $b_{M} = N$, and $\bigoplus_{i=1}^{M} \bm{W}[b_{i-1}\!:\!b_{i}] = \bm{W}$ (i.e. the concatenation of all SUs recovers the input text).

Next, we introduce the SU probability $p_{ _{\mathrm{SU}}} (\bm{W}[i\!:\!j])$ which corresponds to the probability of the text span $\bm{W}[i\!:\!j]$ being an SU.
Based on this probability, the task of sentence segmentation can be formalized as searching for the boundaries $\bm{B}$ which maximize the following probability:\footnote{$M$ is a variable and need not be fixed during the search.}
\begin{equation}
\centering
\argmax_{\bm{B}} \; \prod_{i=1}^{M} \, p_{ _{\mathrm{SU}}} (\bm{W}[b_{i-1}\!:\!b_{i}])
\label{eq:sentence_segmentation_task}
\end{equation}
The most standard approach is to define $p_{ _{\mathrm{SU}}} (\bm{W}[i\!:\!j])$ based on a pretrained EOS labeling model, as we describe in \cref{subsec:sentence_segmentation_method}.
However, our (re-)formulation as Eq. (\ref{eq:sentence_segmentation_task}) is more general and permits other definitions of SU probability as well.

\subsection{Sentence Identification Task}
\label{subsec:sentence_identification_task}

In sentence identification, we consider the input text $\bm{W}$ can be segmented into consecutive, non-overlapping units of SUs and NSUs.
Hence, we regard $\bm{B} = (b_0, b_1,..., b_{M})$ as the unit (SU and NSU) boundaries and define the unit indicators $\bm{A} = (a_1, a_2,..., a_{M})$ for each unit as follows:
\begin{equation*}
\centering 
a_i =
\begin{cases}
    1 & \text{if \: $\bm{W}[b_{i-1}\!:\!b_{i}]$ \: is an SU } \\
    0 & \text{if \: $\bm{W}[b_{i-1}\!:\!b_{i}]$ \: is an NSU } \\
\end{cases}
\end{equation*}
Next, we introduce the NSU probability $p_{ _{\mathrm{NSU}}} (\bm{W}[i\!:\!j])$ which corresponds to the probability of the text span $\bm{W}[i\!:\!j]$ being an NSU.
Based on $p_{ _{\mathrm{SU}}}$ and $p_{ _{\mathrm{NSU}}}$, we can formalize the task of sentence identification as searching for the unit boundaries $\bm{B}$ and unit indicators $\bm{A}$ which maximize the following probability:
\begin{equation}
\centering \resizebox{\hsize}{!}{$
\begin{aligned}
\argmax_{\bm{B}, \bm{A}} \; \prod_{i=1}^{M} \, p_{ _{\mathrm{SU}}} (\bm{W}[b_{i-1}\!:\!b_{i}])^{a_i} \, p_{ _{\mathrm{NSU}}} (\bm{W}[b_{i-1}\!:\!b_{i}])^{1 - a_i}
\end{aligned}
$}
\label{eq:sentence_identification_task}
\end{equation}
Note that this strictly generalizes the sentence segmentation task in Eq. (\ref{eq:sentence_segmentation_task}), which is a special case where $a_i = 1$, $\forall a_i \in \bm{A}$. Based on this task formulation, we discuss how we can define $p_{ _{\mathrm{SU}}} (\bm{W}[i\!:\!j])$ and $p_{ _{\mathrm{NSU}}} (\bm{W}[i\!:\!j])$ to derive our sentence identification algorithm in \cref{subsec:sentence_identification_method}.

\section{Methods}
\label{sec:methods}

\subsection{Sentence Segmentation Method}
\label{subsec:sentence_segmentation_method}

In the most standard approach, sentence segmentation employs an EOS labeling model $p_{ _\mathrm{EOS}}$ to define the SU probability $p_{ _{\mathrm{SU}}}$ in Eq. (\ref{eq:sentence_segmentation_task}).
To be specific, let $p_{ _\mathrm{EOS}} (w_i | \bm{W}; \theta)$ denote the EOS labeling model, which computes the probability of $w_i$ being EOS in $\bm{W}$ ($\theta$ denotes the model parameters).
Typically, it is straightforward to train this model in a \textit{supervised learning} setup using a dataset annotated with gold EOS boundaries \citep{wicks-post-2021-unified}. 
For brevity, we use the notation $p_{ _\mathrm{EOS}} (w_i)$ as a shorthand for $p_{ _\mathrm{EOS}} (w_i | \bm{W}; \theta)$, i.e. we omit $\bm{W}$ and $\theta$ (unless required) in the rest of this paper.

Based on the pretrained model $p_{ _\mathrm{EOS}}$, we can define the SU probability as $p_{ _{\mathrm{SU}}} (\bm{W}[i\!:\!j]) = p_{ _\mathrm{EOS}} (w_{j-1}) \prod_{i \leq k < j-1} (1 - p_{ _\mathrm{EOS}} (w_{k}))$, which requires the last word $w_{j-1}$ to be EOS and all other words to be non-EOS.
By substituting this definition, we can decompose Eq. (\ref{eq:sentence_segmentation_task}) as follows:
\begin{equation}
\centering \resizebox{\hsize}{!}{$
\begin{aligned}
&\text{(\ref{eq:sentence_segmentation_task})} = \argmax_{\bm{B}} \; \sum_{i=1}^{M} \, \log \, p_{ _{\mathrm{SU}}} (\bm{W}[b_{i-1}\!:\!b_{i}]) \\
&= \argmax_{\bm{B}} \; \sum_{i=1}^{M} \, \Big\{\! \log \, p_{ _\mathrm{EOS}} (w_{b_i-1}) \: + \: \sum_{\mathclap{b_{i-1} \leq j < b_i - 1}} \:\, \log \, (1 - p_{ _\mathrm{EOS}} (w_j)) \!\Big\} \\
&= \argmax_{\bm{B}} \sum_{i \in \bm{B}_{\mathrm{EOS}}} \! \log \, p_{ _\mathrm{EOS}} (w_i) \: + \!\! \sum_{\substack{ i \notin \bm{B}_{\mathrm{EOS}}}} \! \log \, (1 - p_{ _\mathrm{EOS}} (w_i)) \\
\end{aligned}
$}
\label{eq:sentence_segmentation_method}
\end{equation}

\noindent
where $\bm{B}_{ _\mathrm{EOS}} = \{b_{i} - 1 \,|\, i \in (1,2,...,M)\}$ represents all the EOS indices defined by $\bm{B}$.

This is a trivial optimization problem where we can simply choose $\bm{B}_{ _\mathrm{EOS}} = \{i \in (0,1,...,N\!-\!1) \,|\, p_{ _\mathrm{EOS}} (w_{i}) \ge 0.5 \}$ to maximize Eq. (\ref{eq:sentence_segmentation_method}). This also shows that sentence segmentation can be conducted by predicting the EOS independently for each $w_i$ based on $p_{ _\mathrm{EOS}} (w_{i})$. In contrast, sentence identification involves a more complex optimization problem which we solve using dynamic programming (\cref{subsec:sentence_identification_method}).

\subsection{Sentence Identification Method}
\label{subsec:sentence_identification_method}

We extend the method of sentence segmentation (\cref{subsec:sentence_segmentation_method}) to conduct sentence identification.
To be specific, we employ pretrained BOS and EOS labeling models $p_{ _\mathrm{BOS}}$, $p_{ _\mathrm{EOS}}$ to define the SU and NSU probabilities $p_{ _{\mathrm{SU}}}$, $p_{ _{\mathrm{NSU}}}$ in Eq. (\ref{eq:sentence_identification_task}).
As a first step, we need to train the BOS and EOS labeling models: this can be conducted in a supervised manner using a dataset containing gold BOS and EOS labels, as we explain in \cref{subsec:model_setup}. 

Based on the pretrained BOS and EOS labeling models, we can define the SU and NSU probabilities as follows:
\begin{equation*}
\centering \resizebox{\hsize}{!}{$
\begin{aligned}
&p_{ _{\mathrm{SU}}} (\bm{W}[i\!:\!j]) = \;\, p_{ _\mathrm{BOS}} (w_i) \prod_{i < k \leq j-1} (1 - p_{ _\mathrm{BOS}} (w_{k})) \;\, \\
& \qquad \qquad \qquad \qquad \times \:\, p_{ _\mathrm{EOS}} (w_{j-1}) \prod_{i \leq k < j-1} (1 - p_{ _\mathrm{EOS}} (w_{k})) \\
& p_{ _{\mathrm{NSU}}} (\bm{W}[i\!:\!j]) = \: \prod_{\mathclap{i \leq k \leq j-1}} \:\, (1 - p_{ _\mathrm{BOS}} (w_{k}))  \,\times\,  \: \prod_{\mathclap{i \leq k \leq j-1}} \:\, (1 - p_{ _\mathrm{EOS}} (w_k)) \\
\end{aligned}
$}
\end{equation*}

In the SU probability $p_{ _{\mathrm{SU}}}$, the first word $w_i$ is required to be BOS, the last word $w_{j-1}$ to be EOS, and all other words to be neither BOS nor EOS. Note that this definition of $p_{ _{\mathrm{SU}}}$ is a natural generalization from \cref{subsec:sentence_segmentation_method} which only relies on the EOS probability $p_{ _\mathrm{EOS}}$.

In contrast, the NSU probability $p_{ _{\mathrm{NSU}}}$ requires all words to be neither BOS nor EOS. Notably, this definition does not distinguish contiguous NSUs in the sense that $p_{ _{\mathrm{NSU}}} (\bm{W}[i\!:\!k]) = p_{ _{\mathrm{NSU}}} (\bm{W}[i\!:\!j]) \times p_{ _{\mathrm{NSU}}} (\bm{W}[j\!:\!k])$ \,if\, $\bm{W}[i\!:\!j] \oplus \bm{W}[j\!:\!k] = \bm{W}[i\!:\!k]$. This is convenient as we are only interested in the extraction of SUs and do not need to seek the exact boundaries between consecutive NSUs.

By substituting these definitions of $p_{ _{\mathrm{SU}}}$ and $p_{ _{\mathrm{NSU}}}$, we can decompose Eq. (\ref{eq:sentence_identification_task}) as follows:
\begin{equation}
\centering \resizebox{\hsize}{!}{$
\begin{aligned}
&\text{(\ref{eq:sentence_identification_task})} = \argmax_{\bm{B}, \bm{A}} \; \sum_{i=1}^{M} \, \Big\{ a_i \log \, p_{ _{\mathrm{SU}}} (\bm{W}[b_{i-1}\!:\!b_{i}]) \\
& \qquad \qquad \qquad \qquad \:\:\:\: + \:\: (1 - a_i) \log \, p_{ _{\mathrm{NSU}}} (\bm{W}[b_{i-1}\!:\!b_{i}]) \Big\} \\
&= \argmax_{\bm{B}, \bm{A}} \:\: \sum_{\mathclap{i \in \bm{B}_{\mathrm{BOS}}^{\bm{A}}}} \: \log \, p_{ _\mathrm{BOS}} (w_i) + \: \sum_{\mathclap{i \notin \bm{B}_{\mathrm{BOS}}^{\bm{A}}}} \: \log \, (1 - p_{ _\mathrm{BOS}} (w_i)) \\
& \qquad \qquad \quad + \:\: \sum_{\mathclap{i \in \bm{B}_{\mathrm{EOS}}^{\bm{A}}}} \log \, p_{ _\mathrm{EOS}} (w_i) + \sum_{\mathclap{i \notin \bm{B}_{\mathrm{EOS}}^{\bm{A}}}} \log \, (1 - p_{ _\mathrm{EOS}} (w_i))
\end{aligned}
$}
\label{eq:sentence_identification_method}
\end{equation}

\noindent
where $\bm{B}_{\scriptscriptstyle \mathrm{BOS}}^{\scriptscriptstyle \bm{A}} = \{b_{i-1} \,|\, i \!\in\! (1,2,...,M), a_i = 1\}$ denotes the BOS indices and $\bm{B}_{\scriptscriptstyle \mathrm{EOS}}^{\scriptscriptstyle \bm{A}} = \{b_{i}-1 \,|\, i \!\in\! (1,2,...,M), a_i = 1\}$ denotes the EOS indices, both defined by $\bm{B}$ and $\bm{A}$.

Therefore, our goal is to choose $\bm{B}_{\scriptscriptstyle \mathrm{BOS}}^{\scriptscriptstyle \bm{A}}$ and $\bm{B}_{\scriptscriptstyle \mathrm{EOS}}^{\bm{\scriptscriptstyle  A}}$ which maximize Eq. (\ref{eq:sentence_identification_method}).
To this end, we need to consider the restrictions that (i) the first label should be BOS, (ii) the last label should be EOS, and (iii) BOS and EOS labels need to appear alternately.
These restrictions can be incorporated in our dynamic programming framework to find the argmax of Eq. (\ref{eq:sentence_identification_method}).
For the precise algorithm, we refer the readers to Appendix \ref{sec:dynamic_programming_algorithm}.

\section{Evaluation}
\label{sec:evaluation}

Due to the novelty of the task, currently there exists no benchmark for evaluating sentence identification.
To address this issue, we propose a fully automatic procedure to convert the Universal Dependencies (UD) corpora \citep{de-marneffe-etal-2021-universal} into sentence identification benchmarks.

Concretely speaking, we conduct the following two steps based on the gold UD annotation: (i) the detection of unit (SU and NSU) boundaries and (ii) the classification of each unit into SU or NSU.
As for (i), we simply use the original \textit{sentence boundaries} in the UD annotation, where UD uses the term \textit{sentence} in a broader sense including both SUs and NSUs (e.g. sentence fragments).
Note that the exact boundaries between consecutive NSUs (which we call NSU--NSU boundaries) do not need to be accurate or consistent, since we are only interested in extracting the spans of SUs.
However, we do expect that the original boundaries are generally reliable in all other cases (SU--SU and SU--NSU boundaries), which seems to be the case.

The main problem is (ii), i.e. how to classify each unit as an SU or NSU.
To this end, we follow the notion of \textit{lexical sentence} in linguistics \citep{nunberg1990linguistics} which defines an SU based on the dependencies among the lexical items, e.g. a group of words that contain a subject and predicate.
In this work, we build upon the UD dependency relations and define an SU as a unit that contains at least one \textit{clausal predicate} with a \textit{core} or \textit{non-core argument}.\footnote{To check this condition, we simply need to verify whether there is at least one core argument (e.g. \textit{nsubj}, \textit{obj}, \textit{ccomp}) or non-core dependent (e.g. \textit{obl}, \textit{advcl}, \textit{aux}). For a full list of the UD relations, see https://universaldependencies.org/u/dep/.}
Here, a \textit{clause} expresses an event or proposition which we regard as an essential aspect of SUs.
A clausal predicate and a core argument form the backbone of a clause, while a non-core argument modifies it \citep{de-marneffe-etal-2021-universal}.

Note that our current definition excludes \textit{noun phrases} appearing by themselves, since they only consist of the \textit{nominal dependent} relations.
However, we can flexibly customize the definition of SUs to include or exclude such phrases.

Due to the reliance on UD, our conversion procedure can be applied to a wide variety of languages supported in UD (currently over 100 languages).
However, as a first set of experiments, we focus on the English Web Treebank (EWT) \citep{silveira-etal-2014-gold} as the primary benchmark of sentence identification.
This dataset comprises five genres of web media texts: namely weblogs, newsgroup threads, emails, product reviews, and Q\&A websites.
Consequently, the dataset contains formal SUs, informal SUs (e.g. without capitalization or punctuations) as well as a wide variety of NSUs.

\begin{table}[t]
\centering \scalebox{0.85}{
\begin{tabular}{l}
\toprule[\heavyrulewidth]
*$\sim$*$\sim$*$\sim$*$\sim$*$\sim$*$\sim$*$\sim$*$\sim$*$\sim$* \\
**********NOTE********** \\
By video conference from \_\_\_\_\_\_\_ \\
Excerpt: \\
02/13/2001 08:02 PM \\
5:00 PT ** 6:00 MT ** 7:00 CT ** 8:00 ET \\
Time: 11:30am / 1:30pm Central / 2:30pm Eastern \\
Sunshine Coast, British Columbia, Canada \\
- UnleadedStocks.pdf \\
t r u t h o u t — Perspective \\
( Answered, 2 Comments ) \\
The federal sites of Washington, DC. \\
From Madrid to Seville to Barcelona an Valencia. \\
\bottomrule[\heavyrulewidth]
\end{tabular}
}
\caption{Examples of gold NSUs in the English Web Treebank (EWT) identified based on our procedure. Each line corresponds to one example of NSU.}
\label{table:nsu_examples}
\end{table}

We show some examples of NSUs in Table \ref{table:nsu_examples} (and more in Appendix \ref{sec:su_and_nsu_examples}) identified based on our procedure.
As shown by the results, our procedure can identify various NSUs including nonlinguistic markers, timestamps, lists, contact information, etc.
We can also see that noun/prepositional phrases are classified as NSUs based on our criteria.
By excluding such NSUs and identifying SUs, we can clearly separate the portions of the text that are worth sophisticated linguistic analyses, e.g. based on dependency parsing or manual inspection.

Finally, we summarize the dataset statistics of EWT in Table \ref{table:ewt_dataset_statistics}.
Overall, 17$\sim$28\% of the units were classified as NSUs, with the test set containing the highest proportion of NSUs.
We also regard SU extraction as a word-level or character-level BIO labeling task and report the number of gold BIO labels in Table \ref{table:ewt_dataset_statistics}.\footnote{B $=$ Beginning of SU, I $=$ Inside of SU, and O $=$ Outside of SU. Details of how we assign the gold BIO labels (at the word-level and character-level) are provided in Appendix \ref{sec:label_assignment_and_conversion}.}
At the word-level, we can see that the proportion of O-labels (indicating NSUs) is only 4$\sim$8\% and much smaller than the proportion of NSUs in terms of units: this is because NSUs are usually short and contain only a few words.
At the character-level, the proportion of O-labels is slightly larger (6$\sim$13\%): this is because NSUs often contain extraordinarily long words like URLs and long sequences of nonlinguistic symbols.

Overall, we could verify that there exists a non-negligible amount of NSUs in the EWT dataset, which we aim to exclude with sentence identification in our experiments.

\begin{table}[t]
\centering \scalebox{0.85}{
\begin{tabular}{crrrr}
\toprule[\heavyrulewidth]
 & & \multicolumn{1}{c}{Train} & \multicolumn{1}{c}{Dev} & \multicolumn{1}{c}{Test} \\
\midrule
\multicolumn{2}{c}{Total SUs} & 10,356 & 1,523 & 1,490 \\
\multicolumn{2}{c}{Total NSUs} & 2,187 & 478 & 587 \\
\midrule
\multirow{3}{*}{\shortstack[c]{Word-\\Level}} & B-Label & 10,356 & 1,523 & 1,490 \\
 & I-Label & 160,127 & 18,791 & 18,222 \\
 & O-Label & 6,939 & 1,302 & 1,822 \\
\midrule
\multirow{3}{*}{\shortstack[c]{Character-\\Level}} & B-Label & 10,356 & 1,523 & 1,490 \\
 & I-Label & 773,223 & 92,309 & 88,441 \\
 & O-Label & 47,107 & 9,925 & 13,232 \\
\bottomrule[\heavyrulewidth]
\end{tabular}
}
\caption{EWT dataset statistics.}
\label{table:ewt_dataset_statistics}
\end{table}

\section{Experimental Setup}
\label{sec:experimental_setup}

\subsection{Model Setup}
\label{subsec:model_setup}

As we discussed in \cref{subsec:sentence_identification_method}, our sentence identification method requires pretrained BOS and EOS labeling models to identify SUs and NSUs.
To develop these models, we simply finetune \roberta by adding a binary BOS/EOS classifier on top of the encoder.

To enable our models to handle various lengths of the input texts, we concatenate the consecutive $L$ units of gold SUs and NSUs as the input during training, where $L$ is sampled from a geometric distribution with parameter $p_{ _{CC}}$.\footnote{With parameter $p_{ _{CC}}\in(0,1]$, the probability mass function of the geometric distribution is \,$p(L=l) \,=\, (1 - p_{ _{CC}})^{l-1} p_{ _{CC}}$ where $l \in \{1, 2, 3,...\}$. As $p_{ _{CC}}$ decreases, the distribution gets more skewed towards larger $L$. With $p_{ _{CC}}=0$, we consider $p(L=\infty) = 1$.}
However, the RoBERTa encoder has the restriction that the input text size cannot exceed 512 subwords.
Therefore, if the input text size is too large, we replace $L$ with the maximum $L^\prime < L$ which satisfies this restriction.
Note that this is a common procedure to sample variable (instead of fixed) lengths of concatenated units \citep{joshi-etal-2020-spanbert}.

Assuming the existence of the in-domain sentence identification dataset (EWT Train/Dev), it is straightforward to train the BOS/EOS labeling models based on our unit concatenation procedure.
However, we may not always have the gold annotation of SUs and NSUs for the target domain.
To take such cases into account, we also consider a setup where we only have the standard sentence segmentation dataset (WSJ Train/Dev) to train the BOS/EOS labeling models.

When using the sentence segmentation dataset (WSJ), we need to apply the unit concatenation procedure using only clean, edited SUs. 
Unfortunately, this can yield the following data priors which do not actually hold in a sentence identification dataset (EWT): (i) an SU (almost) always starts with a capitalization and ends with punctuation, (ii) the first word of the input is always BOS and the last word is always EOS, and (iii) BOS always directly follows EOS.

To address (i) and (ii), we propose a simple data augmentation technique to alleviate the discrepancy in the data priors. To address (iii), we propose an ensembling technique with the unidirectional (instead of bidirectional) models which are agnostic to this data prior.

\subsubsection{Data Augmentation (+AUG)} To address (i), we conduct a unit-level data augmentation, i.e. we modify each unit based on the following rules with a small probability $p_{ _{DA}}$:

\begin{itemize}
 \item Convert all words in the unit to lower-case, upper-case, or title-case (e.g. ``\textit{hello world}'', ``\textit{HELLO WORLD}'', or ``\textit{Hello World}'').
 \item Remove sentence ending punctuations based on a regular-expression matcher (following \textsc{Ersatz}, \citealp{wicks-post-2021-unified}).
\end{itemize}

\noindent
After the unit-level augmentation, we can apply the unit concatenation in the exact same manner.

Finally, to address (ii), we randomly apply a unit truncation to the first and last units of the concatenated input.
To be specific, we choose a random word in the first (last) unit and remove all words prior (posterior) to it with a small probability $p_{ _{TR}}$.
If the truncation is conducted, we regard the unit as an NSU and fix the gold BOS/EOS labels accordingly. See Table \ref{table:data_augmentation} for an illustration.

Based on this procedure, we can expect to alleviate the data priors (i) and (ii). For more details, we refer the readers to Appendix \ref{sec:details_model_setup}.

\begin{table}[t]
\centering \scalebox{0.81}{
\begin{tabular}{cll}
\toprule[\heavyrulewidth]
\multicolumn{2}{c}{\multirow{2}{*}{Orig.}} & \texttt{\textbf{B}}\phantom{\texttt{oe went to school}}\texttt{\textbf{E} }\texttt{\textbf{B}} \\
& & \texttt{\SU{Joe went to school.} }\SU{\texttt{After that he} ...}\\
\midrule
\multirow{2}{*}{(i)\!\!\!} & Unit & \texttt{\textbf{B}}\phantom{\texttt{oe went to schoo}}\texttt{\textbf{E}\phantom{.} }\texttt{\textbf{B}} \\
& Aug. & \texttt{\SU{Joe went to school}\phantom{.} }\SU{\texttt{AFTER THAT HE} ...}\\
\midrule
\multirow{2}{*}{(ii)\!\!\!} & Unit & {\color{lightgray}{\texttt{\textbf{\st{B}}}}}\phantom{\texttt{oe went to schoo}}{\color{lightgray}\texttt{\textbf{\st{E}}\phantom{.} }}\texttt{\textbf{B}} \\
& Trunc. & {\color{lightgray}{\texttt{\st{Joe went}}}}\texttt{ to school\phantom{.} }\SU{\texttt{AFTER THAT HE} ...}\\
\bottomrule[\heavyrulewidth]
\end{tabular}
}
\caption{Illustration of our data augmentation technique.
In (i) \textit{unit-level augmentation}, we randomly change the casing or remove the last punctuations of each unit.
In (ii) \textit{unit truncation}, we randomly truncate the first and last units of the input (and regard them as NSUs).}
\label{table:data_augmentation}
\end{table}

\subsubsection{Unidirectional Model (+UNI)}
\begin{figure}[t]
\centering
\includegraphics[width=0.99\columnwidth]{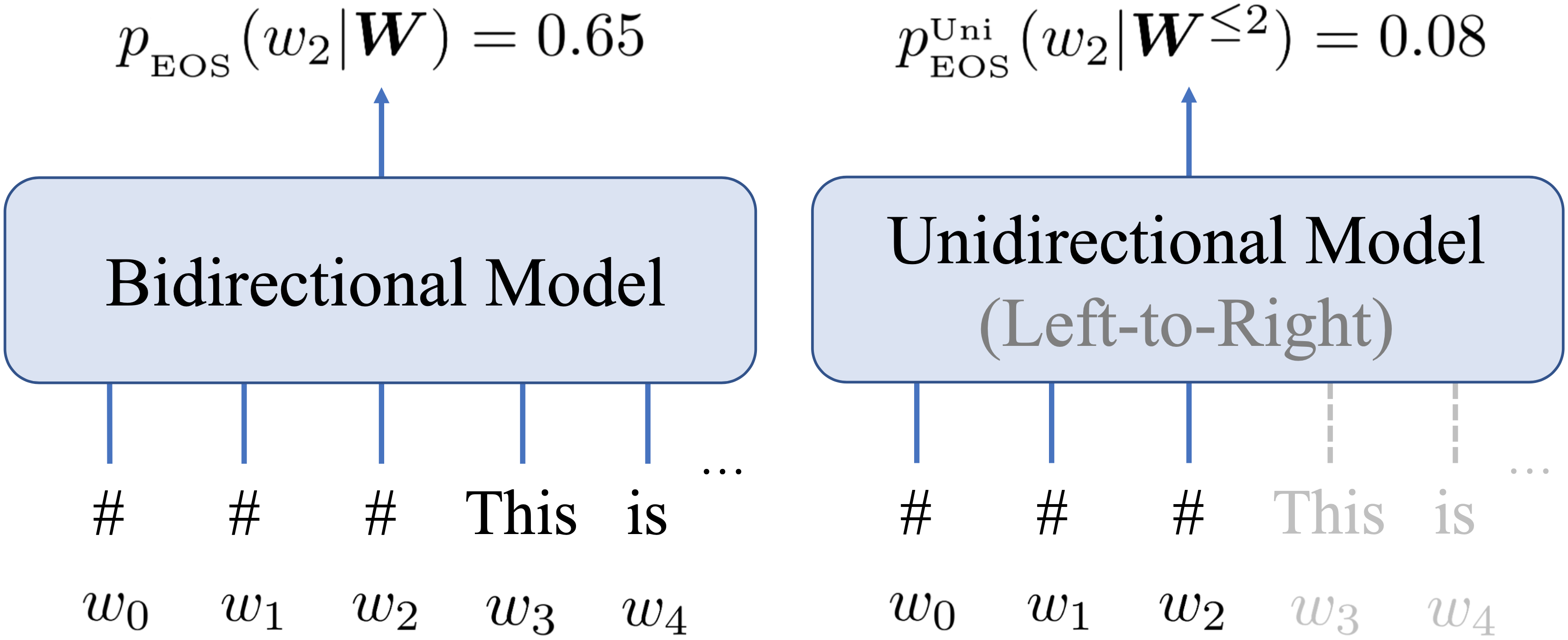}
\caption{Illustration of the bidirectional EOS model (left) and the unidirectional EOS model (right).}
\label{fig:unidirectional_model}
\end{figure}

Simply concatenating SUs (without NSUs) yields the data prior (iii), i.e. BOS always directly follows EOS.
This prior can be easily captured by the bidirectional models $p_{ _\mathrm{BOS}} (w_i | \bm{W})$, $p_{ _\mathrm{EOS}} (w_i | \bm{W})$ conditioned on the whole input $\bm{W}$, including our RoBERTa-based models.
For instance, as shown in Figure \ref{fig:unidirectional_model}, the model may predict EOS at the end of the first unit ($w_2 =$ \textit{\#}) just because the next word ($w_3 =$ \textit{This}) is likely predicted as BOS.

To alleviate this issue, we propose to combine the predictions of the unidirectional models for BOS and EOS labeling. To be precise, let $\bm{W}^{\leq i} = (w_0,...,w_i)$ and $\bm{W}^{\geq i} = (w_i,...,w_{N-1})$. Then, we can represent the unidirectional BOS model as $p^{ _\mathrm{Uni}}_{ _\mathrm{BOS}} (w_i | \bm{W}^{\geq i})$ (looking the context right-to-left) and EOS model as $p^{ _\mathrm{Uni}}_{ _\mathrm{EOS}} (w_i | \bm{W}^{\leq i})$ (looking left-to-right). As illustrated in Figure \ref{fig:unidirectional_model}, these models are agnostic to the data prior (iii). In practice, we can simply use different attention masks and share the encoder parameters (except the last classifier) for the unidirectional and bidirectional models.

We can utilize these unidirectional models by taking a linear intepolation with the bidirectional models as follows:
\begin{equation*}
\centering \resizebox{\hsize}{!}{$
{\footnotesize
\begin{aligned}
p^{ _\mathrm{+Uni}}_{ _\mathrm{BOS}} (w_i | \bm{W}) = \lambda \cdot p^{ _\mathrm{Uni}}_{ _\mathrm{BOS}} (w_i | \bm{W}^{\geq i}) + (1\!-\!\lambda) \cdot p_{ _\mathrm{BOS}} (w_i | \bm{W}) \\
p^{ _\mathrm{+Uni}}_{ _\mathrm{EOS}} (w_i | \bm{W}) = \lambda \cdot p^{ _\mathrm{Uni}}_{ _\mathrm{EOS}} (w_i | \bm{W}^{\leq i}) + (1\!-\!\lambda) \cdot p_{ _\mathrm{EOS}} (w_i | \bm{W}) \\
\end{aligned}
}
$}
\end{equation*}
Then, we can use $p^{ _\mathrm{+Uni}}_{ _\mathrm{BOS}}$ and $p^{ _\mathrm{+Uni}}_{ _\mathrm{EOS}}$ in place of $p_{ _\mathrm{BOS}}$ and $p_{ _\mathrm{EOS}}$ (respectively) to conduct sentence identification, as described in \cref{subsec:sentence_identification_method}.\\

Finally, we compare our proposed methods against sentence segmentation baselines which only utilize EOS labels.\footnote{This EOS-only method is the most reasonable baseline to quantify the precise advantage from combining BOS labels in addition to EOS, which is proposed in our methods.}
As for the baselines, we use the EOS labeling model developed in the same manner to segment the input text based on EOS.
Note that we can optionally force the last word in the input to be EOS: in this case, the result will only contain SUs since all segments will end with EOS.
By default, we do not force the last EOS: in this case, the segment after the last EOS (if exists) is considered as an NSU.

As a default configuration, we use $p_{ _{CC}}\!=\!0.5$, $p_{ _{DA}}\!=\!0.3$, $p_{ _{TR}}\!=\!0.1$, and $\lambda\!=\!0.5$ in our experiments.
To ensure reproducibility, we report more details on the hyperparameters and model setup in Appendix \ref{sec:details_model_setup}.
For the precise procedure on how we convert between the word-, character-, and subword-level labels (for RoBERTa), we refer the readers to Appendix \ref{sec:label_assignment_and_conversion}.

\subsection{Evaluation Setup}
\label{subsec:evaluation_setup}

In the evaluation phase, we consider three ways of assembling the input texts on which we conduct sentence identification.
Firstly, we can apply the same unit concatenation procedure as described in \cref{subsec:model_setup}.
To be specific, we use $p_{ _{CC}}\!=\!0.5$ (same as the training phase) and $p_{ _{CC}}\!=\!0$ (which concatenates the units up to the maximal length) to simulate both shorter and longer lengths of the input texts.

However, this approach is relatively \textit{synthetic} in the sense that we take the gold unit boundaries for granted.
They are usually unavailable at the inference time, so we should consider a more realistic setting for evaluating the methods without relying on the gold unit boundaries.

To this end, we propose to evaluate sentence identification as a \textit{postprocessing} of sentence segmentation.
To be specific, we first apply the state-of-the-art method \textsc{Ersatz} \citep{wicks-post-2021-unified} on the raw text of EWT and then apply sentence identification to each segmented text.
Note that \textsc{Ersatz} has high precision but still predicts false EOS which can fragment a gold SU: in such cases, we consider the fragmented SUs as NSUs and fix the labels accordingly (just as we did in unit truncation, cf. \cref{subsec:model_setup} and Table \ref{table:data_augmentation}).

As for the evaluation metrics, we convert the predictions of our methods into word/character-level BIO labels (cf. Appendix \ref{sec:label_assignment_and_conversion}) and compute the F1 score for each label prediction.
Then, we summarize the results as the macro average F1 and weighted average F1.
We also compute the F1 score of the exact SU span extraction at the word/character-level.
Finally, we run each experiment (from model training to testing) five times with different random seeds and report the average and standard deviation as the final results.

\section{Results}
\label{sec:experiment_results}

\begin{table*}[t]
\centering \scalebox{0.76}{
\setlength\tabcolsep{4.4pt}
\begin{tabular}{llccccccccc}
\toprule[\heavyrulewidth]
\multirow{3}{*}{\shortstack[c]{Train/Dev\\Datasets}} & \multirow{3}{*}{Model} & \multicolumn{3}{c}{EWT Test ($p_{ _{CC}}=0.5$)} & \multicolumn{3}{c}{EWT Test ($p_{ _{CC}}=0$)} & \multicolumn{3}{c}{EWT Test (Postprocess)} \\
\cmidrule(lr){3-5} \cmidrule(lr){6-8} \cmidrule(lr){9-11}
  & & BIO & BIO & \multirow{2}{*}{Span} & BIO & BIO & \multirow{2}{*}{Span} & BIO & BIO & \multirow{2}{*}{Span} \\
 & & Macro & Weighted & & Macro & Weighted & & Macro & Weighted & \\
\midrule
\multirow{3}{*}{\shortstack[c]{EWT\\Train/Dev}} & EOS-Only & 83.2{\scriptsize $\pm$1.5} & 93.9{\scriptsize $\pm$0.6} & 72.8{\scriptsize $\pm$1.8} & 59.7{\scriptsize $\pm$0.2} & 86.4{\scriptsize $\pm$0.1} & 58.2{\scriptsize $\pm$1.1} & 86.3{\scriptsize $\pm$2.7} & 94.6{\scriptsize $\pm$1.1} & 81.6{\scriptsize $\pm$2.4} \\
 & EOS-Only (force last) & 58.6{\scriptsize $\pm$0.1} & 86.6{\scriptsize $\pm$0.0} & 60.4{\scriptsize $\pm$0.8} & 57.6{\scriptsize $\pm$0.2} & 85.9{\scriptsize $\pm$0.1} & 57.7{\scriptsize $\pm$1.0} & 59.1{\scriptsize $\pm$0.1} & 85.7{\scriptsize $\pm$0.0} & 62.3{\scriptsize $\pm$0.3} \\
 & BOS\&EOS & \textbf{93.0{\scriptsize $\pm$1.4}} & \textbf{97.3{\scriptsize $\pm$0.6}} & \textbf{87.3{\scriptsize $\pm$1.6}} & \textbf{91.0{\scriptsize $\pm$1.8}} & \textbf{96.4{\scriptsize $\pm$0.7}} & \textbf{84.1{\scriptsize $\pm$2.6}} & \textbf{92.3{\scriptsize $\pm$1.0}} & \textbf{96.7{\scriptsize $\pm$0.4}} & \textbf{88.8{\scriptsize $\pm$0.9}} \\
\midrule
\multirow{6}{*}{\shortstack[c]{WSJ\\Train/Dev}} & EOS-Only & 71.7{\scriptsize $\pm$0.7} & 88.9{\scriptsize $\pm$0.4} & 59.2{\scriptsize $\pm$2.4} & 56.9{\scriptsize $\pm$0.6} & 85.2{\scriptsize $\pm$0.3} & 48.2{\scriptsize $\pm$2.5} & 71.5{\scriptsize $\pm$0.3} & 87.8{\scriptsize $\pm$0.3} & 67.8{\scriptsize $\pm$0.3} \\
 & EOS-Only (force last) & 57.5{\scriptsize $\pm$0.3} & 86.2{\scriptsize $\pm$0.2} & 53.6{\scriptsize $\pm$2.1} & 55.4{\scriptsize $\pm$0.7} & 85.0{\scriptsize $\pm$0.3} & 48.2{\scriptsize $\pm$2.5} & 58.9{\scriptsize $\pm$0.1} & 85.7{\scriptsize $\pm$0.0} & 61.1{\scriptsize $\pm$0.2} \\
 & EOS-Only (+AUG) & 66.4{\scriptsize $\pm$1.5} & 88.3{\scriptsize $\pm$0.4} & 59.5{\scriptsize $\pm$1.4} & 58.3{\scriptsize $\pm$0.5} & 86.1{\scriptsize $\pm$0.3} & 54.4{\scriptsize $\pm$2.5} & 71.1{\scriptsize $\pm$1.3} & 88.5{\scriptsize $\pm$0.6} & 66.2{\scriptsize $\pm$1.9} \\
 & BOS\&EOS & 71.5{\scriptsize $\pm$0.2} & 89.1{\scriptsize $\pm$0.2} & 59.1{\scriptsize $\pm$1.5} & 57.7{\scriptsize $\pm$0.9} & 85.4{\scriptsize $\pm$0.2} & 48.8{\scriptsize $\pm$1.6} & 71.0{\scriptsize $\pm$0.3} & 87.9{\scriptsize $\pm$0.2} & 68.4{\scriptsize $\pm$0.3} \\
 & BOS\&EOS (+UNI) & 70.4{\scriptsize $\pm$0.7} & 88.2{\scriptsize $\pm$0.3} & 60.0{\scriptsize $\pm$1.1} & 63.3{\scriptsize $\pm$0.8} & 86.0{\scriptsize $\pm$0.4} & 53.0{\scriptsize $\pm$1.3} & 70.8{\scriptsize $\pm$0.4} & 87.6{\scriptsize $\pm$0.2} & 68.4{\scriptsize $\pm$0.1} \\
 & BOS\&EOS (+UNI +AUG) & \textbf{72.5{\scriptsize $\pm$0.4}} & \textbf{89.5{\scriptsize $\pm$0.1}} & \textbf{66.6{\scriptsize $\pm$0.2}} & \textbf{72.4{\scriptsize $\pm$1.3}} & \textbf{89.1{\scriptsize $\pm$0.5}} & \textbf{63.7{\scriptsize $\pm$1.0}} & \textbf{74.3{\scriptsize $\pm$1.1}} & \textbf{89.6{\scriptsize $\pm$0.4}} & \textbf{71.9{\scriptsize $\pm$1.4}} \\
\bottomrule[\heavyrulewidth]
\end{tabular}
}
\caption{\textbf{Overall Results} (Word-Level).
We report the macro/weighted average F1 of the BIO labeling task and the F1 score of the exact SU span extraction task.
Details of our experimental setup are discussed in \cref{sec:experimental_setup}.
}
\label{table:roberta_overall_word_results}
\end{table*}

Table \ref{table:roberta_overall_word_results} summarizes the word-level evaluation results. The results for the character-level evaluation show similar tendencies, so we put them in Appendix \ref{sec:full_experimental_results}.
The F1 score for each BIO label prediction is also available in Appendix \ref{sec:full_experimental_results}.

Firstly, we take a look at the results when we have the in-domain sentence identification dataset (EWT Train/Dev) for model development. In this setup, we can verify that our proposed method (BOS\&EOS) significantly outperforms the baselines (EOS-Only) in all metrics. For instance, our method achieves consistently high performance of 84$\sim$89\% F1 for the exact SU span extraction, both at the word- and character-level. This is a very promising result that demonstrates the effectiveness of our method when we can leverage the gold SUs and NSUs from the target domain.

Secondly, we focus on the results where we only utilize the standard sentence segmentation dataset (WSJ Train/Dev) for model development. In this setup, we also report the results of applying our data augmentation (+AUG) and unidirectional model (+UNI) techniques from \cref{subsec:model_setup}.\footnote{We did not observe any improvement from applying these techniques to the in-domain dataset (EWT Train/Dev), which is consistent with our motivation and expectation.}

Due to the data discrepancy between WSJ and EWT, we find a natural drop in performance compared to the previous setup using in-domain EWT Train/Dev. However, we can verify that our techniques (+AUG, +UNI) generally help to alleviate this issue, and our proposed method performs on par or slightly better than the EOS-only baselines when applying these techniques.
It is especially worth noting the improvement in the exact SU span extraction task (reaching 64$\sim$72\% F1), where the advantage of our method is the most conspicuous and consistent in both word- and character-level evaluation.
This improvement can also be explained by the higher performance in the B-label prediction with our method (Appendix \ref{sec:full_experimental_results}), which is a prerequisite for accurate SU span extraction.

Finally, we note that the EOS-only baseline without forcing the last EOS can be quite competitive with shorter inputs ($p_{ _{CC}}=0.5$ and postprocessing) but performs considerably worse when the input texts are longer ($p_{ _{CC}}=0$). This is because the baseline can only predict the last segment of the input as an NSU, which is less problematic when the input texts are shorter but becomes increasingly problematic with longer inputs (since most NSUs will not be able to be removed). In contrast, our proposed method performs much more robustly under various input lengths.

Through further experiments and analyses, we found that (i) the results are stable across different hyperparameter choices, (ii)
predictions are reasonable especially when using the in-domain dataset (EWT Train/Dev) for model development, and (iii) our methods do not sacrifice performance on the formal/edited texts of the sentence segmentation dataset (WSJ Test). These detailed evidences can be found in Appendix \ref{sec:further_experiments_and_analyses}.

\section{Conclusion}
\label{sec:conclusion}

In this paper, we introduced a novel task of sentence identification, where we aim to identify SUs while excluding NSUs in a given text (\cref{sec:task_formulation}).
Through sentence identification, we can clearly distinguish the portions of the text that are appropriate (or not) for the prediction and evaluation of sophisticated linguistic analyses, such as dependency parsing, semantic role labeling, etc.

To conduct sentence identification, we proposed a simple yet effective method of combining the BOS and EOS labeling models to determine the SUs and NSUs (\cref{sec:methods}). To evaluate sentence identification, we designed an automatic, language-independent procedure to convert the UD corpora into sentence identification benchmarks (\cref{sec:evaluation}).

In our experiments, we developed the BOS/EOS labeling models by finetuning pretrained RoBERTa (\cref{sec:experimental_setup}).
Based on the experimental results, we showed that our proposed method combining the BOS and EOS labels outperforms sentence segmentation baselines which only utilize EOS labels in all of the considered settings (\cref{sec:experiment_results}). Overall, we expect sentence identification to be a fundamental framework for the preprocessing of noisy, informal, or non-standard texts in the real world.

\section*{Limitations}
\label{sec:limitations}

Firstly, our current experiments are limited to English and cover only five domains of web media texts in EWT.
However, our task formulation (\cref{sec:task_formulation}), method (\cref{sec:methods}), and evaluation framework (\cref{sec:evaluation}) are fully agnostic to the language and domain.
Hence it is straightforward to conduct experiments in different languages or domains (as long as they are supported in the UD).
While we expect similar results with different languages/domains, we leave further investigation as a future work.

Secondly, while our method performs reliably when the in-domain dataset is available, there is still a huge room left for improvement without relying on such resources (e.g. only using the standard sentence segmentation dataset).
To make our method fully practical, we still need to improve on the accuracy and robustness in such cross-domain scenarios.
One potential approach is to refine the definitions of SU and NSU probabilities from \cref{subsec:sentence_identification_method} to make sentence identification more robust.
For instance, we can incorporate span-level scores instead of only using word-level BOS/EOS probabilities to define the SU/NSU probabilities.
We leave further improvement and extension of our approach as an important future work.

Finally, our methods are currently evaluated on the (exact) SU span extraction task.
Ideally, we should also evaluate the methods on downstream applications such as POS tagging, syntactic parsing, semantic role labeling, etc.
However, we still expect that the (exact) SU span extraction will play a primary role in the evaluation, since accurate (say human-level) identification of SUs/NSUs will likely provide unprecedented benefits on a wide variety of NLP applications dealing with real-world texts.
While we leave the precise analyses on downstream applications as future work, our contributions make the first foundational step towards expanding the capability of the long-established sentence segmentation task.

\newpage

\bibliography{custom}
\bibliographystyle{acl_natbib}

\appendix

\newpage
\section{Dynamic Programming Algorithm}
\label{sec:dynamic_programming_algorithm}

\begin{figure*}[t]
\centering
\includegraphics[width=0.99\textwidth]{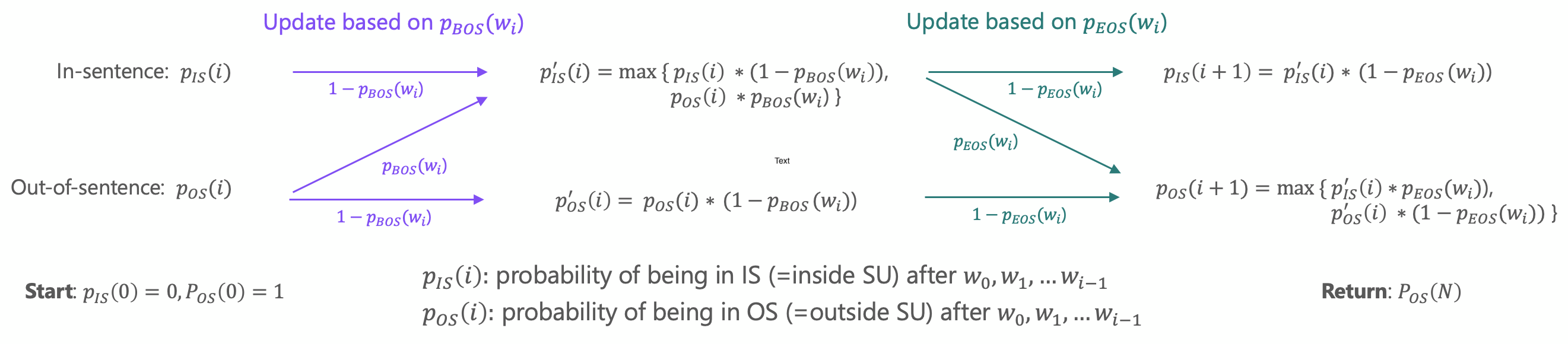}
\caption{Illustration of the dynamic programming procedure.}
\label{fig:dynamic_programming}
\end{figure*}

To find the maximum value (and the argmax) of Eq. \ref{eq:sentence_identification_method} from \cref{subsec:sentence_identification_method}, we rely on a simple dynamic programming framework.
To be specific, we consider the partial labeling of BOS and EOS up to $\bm{W}^{\leq k} = (w_0,...,w_k)$, where $k \leq N-1$.
Then, we aim to compute the maximum log probability of Eq. \ref{eq:sentence_identification_method} based on the partial labeling, i.e. using $\bm{W}^{\leq k}$ in place of $\bm{W}$.

Since the labeling is partial, $\bm{W}^{\leq k}$ may end inside the SU (i.e. the last label is BOS) or outside the SU (i.e. the last label is EOS).
Let $\log \, p_{ _\mathrm{IS}}(k+1)$ denote the maximum log probability when $\bm{W}^{\leq k}$ ends inside the SU and $\log \, p_{ _\mathrm{OS}}(k+1)$ the maximum log probability when $\bm{W}^{\leq k}$ ends outside the SU.
Then, we can initialize $\log \, p_{ _\mathrm{IS}}(0) = \log \, 0 = - \infty$, $\log \, p_{ _\mathrm{OS}}(0) = \log \, 1 = 0$ (since we always start outside the SU) and iteratively update the two values as follows:
\begin{equation}
\centering \resizebox{\hsize}{!}{$
\begin{aligned}
&\log \, p_{_\mathrm{IS}}^\prime(i) \,=\, \max \, \{ \, \log \, p_{ _\mathrm{IS}}(i) \,+\, \log \, (1\!-\!p_{ _\mathrm{BOS}}(w_i)), \\
&\phantom{\log \, p_{_\mathrm{IS}}^\prime(i) \,=\, \max \, \{ \, } \log \, p_{ _\mathrm{OS}}(i) \,+\, \log \, p_{ _\mathrm{BOS}}(w_i) \, \}\\
&\log \, p_{_\mathrm{OS}}^\prime(i) \,=\, \log \, p_{ _\mathrm{OS}}(i) \,+\, \log \, (1\!-\!p_{ _\mathrm{BOS}}(w_i)) \\
&\log \, p_{_\mathrm{IS}}(i\!+\!1) \,=\, \log \, p_{_\mathrm{IS}}^\prime(i) \,+\, \log \, (1\!-\!p_{ _\mathrm{EOS}}(w_i)) \\
&\log \, p_{_\mathrm{OS}}(i\!+\!1) \,=\, \max \, \{ \, \log \, p_{ _\mathrm{IS}}^\prime (i) \,+\, \log \, p_{ _\mathrm{EOS}} (w_i),\\
&\phantom{\log \, p_{_\mathrm{OS}}(i\!+\!1) \,=\, \max \, \{ \, } \log \, p_{ _\mathrm{OS}}^\prime (i) \,+\, \log \, (1\!-\!p_{ _\mathrm{EOS}}\!(w_i)) \, \}
\end{aligned}
$}
\label{eq:dynamic_programming}
\end{equation}

\noindent
Note that we first update $p_{ _\mathrm{IS}}(i) \!\rightarrow\! p_{ _\mathrm{IS}}^\prime(i)$ and $p_{ _\mathrm{OS}}(i) \!\rightarrow\! p_{ _\mathrm{OS}}^\prime(i)$ based on the BOS probability $p_{ _\mathrm{BOS}}(w_i)$.
Then, we update $p_{ _\mathrm{IS}}^\prime(i) \!\rightarrow\! p_{ _\mathrm{IS}}(i\!+\!1)$ and $p_{ _\mathrm{OS}}^\prime(i) \!\rightarrow\! p_{ _\mathrm{OS}}(i\!+\!1)$ based on the EOS probability $p_{ _\mathrm{EOS}}(w_i)$.\footnote{Note that if a single word $w_i$ is labeled as both BOS and EOS at the same time, we can extract it as a single SU.}
The iterative procedure is illustrated in Figure \ref{fig:dynamic_programming}.

Finally, we can compute the log probability $\log \, p_{ _\mathrm{OS}}(N)$ (since we always end outside the SU), which corresponds to the maximum value of Eq. \ref{eq:sentence_identification_method}.
To obtain the argmax, we can simply incorporate backtracking during the iterative updates of Eq. \ref{eq:dynamic_programming}.
Through this dynamic programming framework, we can ensure that the restrictions from \cref{subsec:sentence_identification_method} are satisfied: namely, (i) the first label should be BOS, (ii) the last label should be EOS, and (iii) BOS and EOS labels need to appear alternately.

In practice, we can limit the candidates of BOS indices to the subset where $p_{ _\mathrm{BOS}}(w_i)$ is higher than a certain threshold $c$.
This can be efficiently implemented by simply skipping the updates of $p_{_\mathrm{IS}}^\prime(i)$ and $p_{_\mathrm{OS}}^\prime(i)$, i.e. using $p_{_\mathrm{IS}}^\prime(i) = p_{_\mathrm{IS}}(i)$ and $p_{_\mathrm{OS}}^\prime(i) = p_{_\mathrm{OS}}(i)$, if $p_{ _\mathrm{BOS}}(w_i) < c$.\footnote{This is equivalent to forcing $w_i$ to be non-BOS, i.e. setting $p_{ _\mathrm{BOS}}(w_i) = 0$ in Eq. \ref{eq:dynamic_programming}.}
Likewise, we can limit the candidates of EOS indices by skipping the updates of $p_{_\mathrm{IS}}(i\!+\!1)$ and $p_{_\mathrm{OS}}(i\!+\!1)$ if $p_{ _\mathrm{EOS}}(w_i) < c$.
Generally speaking, this leads to a more efficient algorithm: therefore, we use the candidate threshold of $c = 0.1$ for restricting both BOS and EOS indices throughout our experiments.

\section{SU and NSU Examples}
\label{sec:su_and_nsu_examples}

\begin{table*}[t]
\centering \scalebox{0.76}{
\begin{tabular}{cl}
\toprule[\heavyrulewidth]
\multirow{10}{*}{SUs} & President Bush on Tuesday nominated two individuals to replace retiring jurists on federal courts in the Washington area. \\
& Unfortunately, Mr. Lay will be in San Jose, CA participating in a conference, where he is a speaker, on June 14. \\
& “In 1972, there was an enormous glut of pilots,” Campenni says. \\
& PS -- There is a happy hour tonight at Scudeiros on Dallas Street (just west of the Met Garage) beginning around 5:00. \\
& 2) Your vet would not prescribe them if they didn't think it would be helpful. \\
& BUT EVERYONE HAS THERE OWN WAY!!!!!! \\
& The motel is very well maintained, and the managers are so accomodating, it's kind of like visiting family each year! ;-) \\
& where can I find the best tours to the Mekong Delta at reasonable prices? \\
& it seems like its healthier too, but its prolly not. \\
& I have wifi at my house, but thats just at my house...is there anyway i can buy some card to make the ipod itself have wifi? \\
\midrule
\multirow{10}{*}{NSUs} & ----{\textgreater}===\}*\{==={\textless}---- \\
& - Lisa\_coverletter.doc {\textless}{\textless} File: Lisa\_coverletter.doc {\textgreater}{\textgreater} \\
& Thur. Sept. 28 - Paris (Versailles or Fontainbleu - half day side trip) \\
& 9.3m - Number of US unemployed in April 2004. \\
& Game 1: Monday, May 28 @ 2:00PM vs. Los Angeles SPARKS \\
& Mixed Tempura.....................8.25 Shrimp or vegetable tempura \& salad. \\
& Infinity stereo, bucket seats, nerf bars, tool box, bed liner, camper tow package, 5 speed manual. \\
& printing, printing, copies, printing, copies, printing, \\
& A++++ !!!!!!!!!!!!!!!!!!!!!!!!!!!!!!!! \\
& Dear Sir / Madam, \\
\bottomrule[\heavyrulewidth]
\end{tabular}
}
\caption{Examples of gold SUs and NSUs in the English Web Treebank (EWT) identified based on our procedure (\cref{sec:evaluation}).
Each line corresponds to one example of SU or NSU.}
\label{table:su_nsu_examples}
\end{table*}

In Table \ref{table:su_nsu_examples}, we provide more examples of SUs and NSUs identified based on our procedure described in \cref{sec:evaluation}.
As for the SUs, we can verify that EWT contains clean, formal SUs with appropriate capitalization and punctuation.
We can also verify that EWT contains various types of \textit{informal} SUs, e.g. that lack capitalization/punctuation, use non-standard casing, end with emoticons, include spelling errors, concatenate consecutive SUs without a space, etc.

\section{Label Assignment and Conversion}
\label{sec:label_assignment_and_conversion}

In this section, we explain the precise procedure on how we (i) assign the gold character-level labels, (ii) convert the character-level labels to word/subword-level labels, and (iii) convert the subword-level labels to character/word-level labels.
We limit our explanation to BIO labels, since it is straightforward to convert them to the combination of BOS and EOS labels (and vice versa).

Firstly, we can assign the gold character-level labels from the UD annotation by taking the character-level alignment, which determines the exact spans of SUs and NSUs.
From the character-level labels, we can assign the word- or subword-level labels based on the following rule:

\begin{itemize}
 \item If the word (or subword) contains a character with the B-label, assign it the B-label.
 \item Else if it contains a character with the I-label, assign the I-label.
 \item Otherwise assign the O-label.
\end{itemize}

\noindent
For instance, this procedure is used to create the subword-level labels for training our BOS/EOS labeling models.

To evaluate our methods, we need to convert the subword-level labels produced by our methods into the character-level labels, which can then be converted into the word-level labels (based on the previous procedure). To convert a subword-level label into a sequence of character-level labels, we apply the following rule (where $n$ denotes the number of characters in the subword):

\begin{itemize}
 \item If the subword has the B-label, the character-level labels are $1$ B-label followed by $n-1$ I-labels.
 \item If the subword has the I-label, the character-level labels are $n$ I-labels.
 \item If the subword has the O-label, the character-level labels are $n$ O-labels.
\end{itemize}

\section{Details on the Model Setup}
\label{sec:details_model_setup}

As discussed in \cref{subsec:model_setup}, we finetune the pretrained \roberta publicly available on the Hugging-Face model hub\footnote{https://huggingface.co/models}.
We add a binary BOS/EOS classifier on top of the encoder, which is a single-layer MLP with a hidden size of 768.
We share the encoder parameters and use different classifiers for the BOS/EOS predictions.
The BOS/EOS models are trained jointly by summing their losses.

When we combine the unidirectional models (+UNI), we take the same approach and use different classifiers for the unidirectional/bidirectional models. 
Again, the encoder parameters are shared and all models are trained jointly.

As for the training data preparation, we apply the unit concatenation and data augmentation (+AUG) \textit{on the fly}, i.e. we see different concatenation and augmentation of the units in each iteration.
The same procedure is applied on the validation set.

During data augmentation, we remove the last sentence ending punctuation based on the following regular-expression, similar to the candidate boundary detector in \textsc{Ersatz} \citep{wicks-post-2021-unified}:

\begin{itemize}
\item $(.^*P_e P^*)$ where $P$ denotes the set of punctuations and $P_e \subset P$ denotes the sentence ending punctuations.
\end{itemize}

\noindent
Since our experiments are conducted on English, we use $P =$ \texttt{\{.?!")'\}} and $P_e =$ \texttt{\{.?!\}}.

Finally, all models are implemented in Pytorch and trained on a single Tesla V100-SXM2-32GB GPU. We use a batch size of 8, accumulate the gradients for 32 batches, and apply the gradient clipping at 1.0 before updating the model weights. As for the optimizer, we use Adam with the initial learning rate of 0.0001 and exponentially decay the learning rate by $\gamma = 0.95$ after each epoch.
We check the validation loss every 200 batches and stop the training early if there is no improvement for 5 consecutive evaluations.

\section{The Full Experimental Results}
\label{sec:full_experimental_results}

In this section, we report the full results of our experiments which did not fit in \cref{sec:experiment_results}. Table \ref{table:roberta_bio_word_results} shows the word-level F1 scores for each B-, I-, and O-label prediction. Table \ref{table:roberta_overall_char_results} shows the overall results for the character-level evaluation.

Generally speaking, we can confirm the same results as observed in \cref{sec:experiment_results}. Firstly, our proposed method significantly outperforms the baselines when we use the EWT Train/Dev dataset for model development. Secondly, our method performs slightly better than (or at least on par with) the baselines when developed on the WSJ Train/Dev dataset. Finally, the baseline without forcing the last EOS is competitive with shorter inputs ($p_{ _{CC}}=0.5$ and postprocessing) but performs considerably worse when the input texts are longer ($p_{ _{CC}}=0$).

\begin{table*}[t]
\centering \scalebox{0.76}{
\setlength\tabcolsep{4.4pt}
\begin{tabular}{llrrrrrrrrr}
\toprule[\heavyrulewidth]
\multirow{2}{*}{\shortstack[c]{Train/Dev\\Datasets}} & \multirow{2}{*}{Model} & \multicolumn{3}{c}{EWT Test ($p_{ _{CC}}=0.5$)} & \multicolumn{3}{c}{EWT Test ($p_{ _{CC}}=0.0$)} & \multicolumn{3}{c}{EWT Test (Postprocess)} \\
\cmidrule(lr){3-5} \cmidrule(lr){6-8} \cmidrule(lr){9-11}
  & & \multicolumn{1}{c}{B-Label} & \multicolumn{1}{c}{I-Label} & \multicolumn{1}{c}{O-Label} & \multicolumn{1}{c}{B-Label} & \multicolumn{1}{c}{I-Label} & \multicolumn{1}{c}{O-Label} & \multicolumn{1}{c}{B-Label} & \multicolumn{1}{c}{I-Label} & \multicolumn{1}{c}{O-Label} \\
\midrule
\multirow{3}{*}{\shortstack[c]{EWT\\Train/Dev}} & EOS-Only & 85.6{\scriptsize $\pm$0.8} & 97.3{\scriptsize $\pm$0.3} & 66.6{\scriptsize $\pm$3.5} & 78.0{\scriptsize $\pm$0.6} & 95.1{\scriptsize $\pm$0.1} & 6.0{\scriptsize $\pm$0.4} & 90.2{\scriptsize $\pm$1.2} & 97.5{\scriptsize $\pm$0.5} & 71.3{\scriptsize $\pm$6.6} \\
 & EOS-Only (force last) & 79.8{\scriptsize $\pm$0.2} & 95.9{\scriptsize $\pm$0.0} & 0.0{\scriptsize $\pm$0.0} & 77.8{\scriptsize $\pm$0.6} & 95.1{\scriptsize $\pm$0.1} & 0.0{\scriptsize $\pm$0.0} & 81.7{\scriptsize $\pm$0.2} & 95.7{\scriptsize $\pm$0.0} & 0.0{\scriptsize $\pm$0.0} \\
 & BOS\&EOS & \textbf{94.3{\scriptsize $\pm$0.6}} & \textbf{98.7{\scriptsize $\pm$0.3}} & \textbf{86.1{\scriptsize $\pm$3.5}} & \textbf{93.0{\scriptsize $\pm$1.0}} & \textbf{98.2{\scriptsize $\pm$0.4}} & \textbf{81.7{\scriptsize $\pm$4.0}} & \textbf{94.7{\scriptsize $\pm$0.3}} & \textbf{98.4{\scriptsize $\pm$0.2}} & \textbf{83.9{\scriptsize $\pm$2.4}} \\
\midrule
\multirow{6}{*}{\shortstack[c]{WSJ\\Train/Dev}} & EOS-Only & 78.7{\scriptsize $\pm$1.3} & 94.4{\scriptsize $\pm$0.4} & \textbf{42.1{\scriptsize $\pm$1.3}} & 71.8{\scriptsize $\pm$1.9} & 94.3{\scriptsize $\pm$0.2} & 4.5{\scriptsize $\pm$0.4} & 83.3{\scriptsize $\pm$0.2} & 93.8{\scriptsize $\pm$0.3} & 37.3{\scriptsize $\pm$0.9} \\
 & EOS-Only (force last) & 76.7{\scriptsize $\pm$0.9} & \textbf{95.6{\scriptsize $\pm$0.1}} & 0.0{\scriptsize $\pm$0.0} & 71.7{\scriptsize $\pm$1.9} & \textbf{94.6{\scriptsize $\pm$0.2}} & 0.0{\scriptsize $\pm$0.0} & 81.0{\scriptsize $\pm$0.4} & \textbf{95.6{\scriptsize $\pm$0.0}} & 0.0{\scriptsize $\pm$0.0} \\
 & EOS-Only (+AUG) & 79.4{\scriptsize $\pm$1.0} & 95.4{\scriptsize $\pm$0.2} & 24.5{\scriptsize $\pm$4.1} & 78.1{\scriptsize $\pm$1.8} & 93.3{\scriptsize $\pm$0.2} & 1.4{\scriptsize $\pm$1.1} & 82.7{\scriptsize $\pm$1.1} & 94.9{\scriptsize $\pm$0.5} & 35.6{\scriptsize $\pm$3.2} \\
 & BOS\&EOS & 79.4{\scriptsize $\pm$0.9} & 94.8{\scriptsize $\pm$0.2} & 40.5{\scriptsize $\pm$0.6} & 72.9{\scriptsize $\pm$1.2} & 94.3{\scriptsize $\pm$0.2} & 5.8{\scriptsize $\pm$2.0} & 83.9{\scriptsize $\pm$0.2} & 94.1{\scriptsize $\pm$0.2} & 35.2{\scriptsize $\pm$0.9} \\
 & BOS\&EOS (+UNI) & 79.8{\scriptsize $\pm$0.6} & 93.9{\scriptsize $\pm$0.2} & 37.5{\scriptsize $\pm$1.5} & 76.2{\scriptsize $\pm$1.3} & 93.3{\scriptsize $\pm$0.3} & 20.2{\scriptsize $\pm$1.2} & 83.8{\scriptsize $\pm$0.1} & 93.8{\scriptsize $\pm$0.1} & 34.7{\scriptsize $\pm$0.9} \\
 & BOS\&EOS (+UNI +AUG) & \textbf{83.7{\scriptsize $\pm$0.1}} & 95.0{\scriptsize $\pm$0.2} & 38.7{\scriptsize $\pm$1.2} & \textbf{83.0{\scriptsize $\pm$0.6}} & \textbf{94.6{\scriptsize $\pm$0.3}} & \textbf{39.7{\scriptsize $\pm$3.5}} & \textbf{85.9{\scriptsize $\pm$0.6}} & 95.2{\scriptsize $\pm$0.2} & \textbf{41.9{\scriptsize $\pm$2.8}} \\
\bottomrule[\heavyrulewidth]
\end{tabular}
}
\caption{\textbf{BIO Labeling Results} (Word-Level). We report the F1 scores for each B-, I- and O-label prediction.}
\label{table:roberta_bio_word_results}
\end{table*}

\begin{table*}[t]
\centering \scalebox{0.76}{
\setlength\tabcolsep{4.4pt}
\begin{tabular}{llccccccccc}
\toprule[\heavyrulewidth]
\multirow{3}{*}{\shortstack[c]{Train/Dev\\Datasets}} & \multirow{3}{*}{Model} & \multicolumn{3}{c}{EWT Test ($p_{ _{CC}}=0.5$)} & \multicolumn{3}{c}{EWT Test ($p_{ _{CC}}=0.0$)} & \multicolumn{3}{c}{EWT Test (Postprocess)} \\
\cmidrule(lr){3-5} \cmidrule(lr){6-8} \cmidrule(lr){9-11}
  & & BIO & BIO & \multirow{2}{*}{Span} & BIO & BIO & \multirow{2}{*}{Span} & BIO & BIO & \multirow{2}{*}{Span} \\
 & & Macro & Weighted & & Macro & Weighted & & Macro & Weighted & \\
\midrule
\multirow{3}{*}{\shortstack[c]{EWT\\Train/Dev}} & EOS-Only & 83.8{\scriptsize $\pm$1.1} & 92.7{\scriptsize $\pm$0.5} & 72.8{\scriptsize $\pm$1.8} & 58.5{\scriptsize $\pm$0.2} & 81.5{\scriptsize $\pm$0.0} & 58.2{\scriptsize $\pm$1.1} & 87.7{\scriptsize $\pm$2.3} & 93.9{\scriptsize $\pm$1.2} & 81.6{\scriptsize $\pm$2.4} \\
 & EOS-Only (force last) & 57.7{\scriptsize $\pm$0.1} & 81.0{\scriptsize $\pm$0.0} & 60.4{\scriptsize $\pm$0.8} & 56.9{\scriptsize $\pm$0.2} & 80.9{\scriptsize $\pm$0.0} & 57.7{\scriptsize $\pm$1.0} & 58.1{\scriptsize $\pm$0.1} & 79.9{\scriptsize $\pm$0.0} & 62.3{\scriptsize $\pm$0.3} \\
 & BOS\&EOS & \textbf{94.0{\scriptsize $\pm$1.0}} & \textbf{97.2{\scriptsize $\pm$0.6}} & \textbf{87.3{\scriptsize $\pm$1.6}} & \textbf{92.2{\scriptsize $\pm$1.5}} & \textbf{96.3{\scriptsize $\pm$0.7}} & \textbf{84.1{\scriptsize $\pm$2.6}} & \textbf{93.5{\scriptsize $\pm$0.6}} & \textbf{96.6{\scriptsize $\pm$0.4}} & \textbf{88.9{\scriptsize $\pm$0.8}} \\
\midrule
\multirow{6}{*}{\shortstack[c]{WSJ\\Train/Dev}} & EOS-Only & \textbf{72.8{\scriptsize $\pm$0.6}} & 86.9{\scriptsize $\pm$0.4} & 59.1{\scriptsize $\pm$2.3} & 56.0{\scriptsize $\pm$0.6} & 80.9{\scriptsize $\pm$0.1} & 48.2{\scriptsize $\pm$2.5} & \textbf{73.3{\scriptsize $\pm$0.4}} & 85.6{\scriptsize $\pm$0.2} & 67.7{\scriptsize $\pm$0.4} \\
 & EOS-Only (force last) & 56.6{\scriptsize $\pm$0.3} & 80.9{\scriptsize $\pm$0.0} & 53.5{\scriptsize $\pm$2.0} & 54.9{\scriptsize $\pm$0.6} & 80.7{\scriptsize $\pm$0.1} & 48.2{\scriptsize $\pm$2.5} & 57.8{\scriptsize $\pm$0.2} & 79.9{\scriptsize $\pm$0.0} & 61.0{\scriptsize $\pm$0.3} \\
 & EOS-Only (+AUG) & 64.3{\scriptsize $\pm$1.5} & 83.5{\scriptsize $\pm$0.6} & 59.5{\scriptsize $\pm$1.4} & 57.4{\scriptsize $\pm$0.5} & 81.0{\scriptsize $\pm$0.2} & 54.4{\scriptsize $\pm$2.5} & 69.2{\scriptsize $\pm$1.7} & 84.0{\scriptsize $\pm$0.8} & 66.2{\scriptsize $\pm$1.9} \\
 & BOS\&EOS & 72.7{\scriptsize $\pm$0.7} & \textbf{87.1{\scriptsize $\pm$0.2}} & 59.1{\scriptsize $\pm$1.5} & 57.8{\scriptsize $\pm$1.9} & 81.6{\scriptsize $\pm$0.7} & 48.8{\scriptsize $\pm$1.6} & 72.4{\scriptsize $\pm$1.0} & 85.2{\scriptsize $\pm$0.5} & 68.3{\scriptsize $\pm$0.3} \\
 & BOS\&EOS (+UNI) & 72.4{\scriptsize $\pm$0.6} & 86.3{\scriptsize $\pm$0.3} & 59.6{\scriptsize $\pm$1.0} & 65.3{\scriptsize $\pm$1.0} & 83.6{\scriptsize $\pm$0.5} & 52.9{\scriptsize $\pm$1.3} & 72.8{\scriptsize $\pm$0.4} & 85.3{\scriptsize $\pm$0.2} & 68.0{\scriptsize $\pm$0.2} \\
 & BOS\&EOS (+UNI +AUG) & 72.2{\scriptsize $\pm$1.3} & 86.1{\scriptsize $\pm$0.6} & \textbf{66.5{\scriptsize $\pm$0.3}} & \textbf{72.8{\scriptsize $\pm$1.8}} & \textbf{86.5{\scriptsize $\pm$0.9}} & \textbf{63.6{\scriptsize $\pm$1.0}} & 73.2{\scriptsize $\pm$1.9} & \textbf{85.7{\scriptsize $\pm$0.9}} & \textbf{71.8{\scriptsize $\pm$1.5}} \\
\bottomrule[\heavyrulewidth]
\end{tabular}
}
\caption{\textbf{Overall Results} (Character-Level). We report the macro/weighted average F1 of the BIO labeling task and the F1 score of the exact SU span extraction task.}
\label{table:roberta_overall_char_results}
\end{table*}

\section{Further Experiments and Analyses}
\label{sec:further_experiments_and_analyses}

In this section, we provide further experiments and analyses to complement our study.
To be specific, we provide discussions on the effect of the choice of hyperparameters (F.1), qualitative analyses based on example model outputs (F.2), and evaluation of sentence identification based on the sentence segmentation dataset (F.3).

\subsection{Effect of Hyperparameters}
\label{subsec:effect_hyperparameters}

\begin{table*}[t]
\centering \scalebox{0.76}{
\setlength\tabcolsep{4.4pt}
\begin{tabular}{llccccccccc}
\toprule[\heavyrulewidth]
\multirow{3}{*}{Evaluation} & \multirow{3}{*}{Augmentation Rates} & \multicolumn{3}{c}{EWT Test ($p_{ _{CC}}=0.5$)} & \multicolumn{3}{c}{EWT Test ($p_{ _{CC}}=0$)} & \multicolumn{3}{c}{EWT Test (Postprocess)} \\
\cmidrule(lr){3-5} \cmidrule(lr){6-8} \cmidrule(lr){9-11}
& & BIO & BIO & \multirow{2}{*}{Span} & BIO & BIO & \multirow{2}{*}{Span} & BIO & BIO & \multirow{2}{*}{Span} \\
& & Macro & Weighted & & Macro & Weighted & & Macro & Weighted & \\
\midrule
\multirow{3}{*}{\shortstack[c]{Word-Level}} & $p_{ _{DA}}\!=\!0.15$, $p_{ _{TR}}\!=\!0.05$ & 71.3{\scriptsize $\pm$1.1} & 89.0{\scriptsize $\pm$0.5} & 65.7{\scriptsize $\pm$1.3} & 71.5{\scriptsize $\pm$0.9} & 88.6{\scriptsize $\pm$0.5} & 62.3{\scriptsize $\pm$1.7} & 73.5{\scriptsize $\pm$1.4} & 89.2{\scriptsize $\pm$0.6} & 71.2{\scriptsize $\pm$1.8} \\
& $p_{ _{DA}}\!=\!0.3$,\phantom{0} $p_{ _{TR}}\!=\!0.1$ & 72.5{\scriptsize $\pm$0.4} & 89.5{\scriptsize $\pm$0.1} & 66.6{\scriptsize $\pm$0.2} & 72.4{\scriptsize $\pm$1.3} & 89.1{\scriptsize $\pm$0.5} & 63.7{\scriptsize $\pm$1.0} & 74.3{\scriptsize $\pm$1.1} & 89.6{\scriptsize $\pm$0.4} & 71.9{\scriptsize $\pm$1.4} \\
& $p_{ _{DA}}\!=\!0.45$, $p_{ _{TR}}\!=\!0.15$ & \textbf{73.2{\scriptsize $\pm$1.0}} & \textbf{90.0{\scriptsize $\pm$0.1}} & \textbf{67.3{\scriptsize $\pm$0.8}} & \textbf{73.0{\scriptsize $\pm$0.9}} & \textbf{89.5{\scriptsize $\pm$0.6}} & \textbf{64.0{\scriptsize $\pm$1.8}} & \textbf{75.1{\scriptsize $\pm$1.3}} & \textbf{90.0{\scriptsize $\pm$0.4}} & \textbf{72.1{\scriptsize $\pm$0.7}} \\
\midrule
\multirow{3}{*}{\shortstack[c]{Character-\\Level}} & $p_{ _{DA}}\!=\!0.15$, $p_{ _{TR}}\!=\!0.05$ & \textbf{72.3{\scriptsize $\pm$1.9}} & \textbf{86.3{\scriptsize $\pm$1.0}} & 65.4{\scriptsize $\pm$1.4} & \textbf{73.6{\scriptsize $\pm$0.7}} & \textbf{86.8{\scriptsize $\pm$0.3}} & 62.2{\scriptsize $\pm$1.7} & \textbf{73.8{\scriptsize $\pm$1.5}} & \textbf{86.1{\scriptsize $\pm$0.8}} & 71.0{\scriptsize $\pm$1.8} \\
& $p_{ _{DA}}\!=\!0.3$,\phantom{0} $p_{ _{TR}}\!=\!0.1$ & 72.2{\scriptsize $\pm$1.3} & 86.1{\scriptsize $\pm$0.6} & 66.5{\scriptsize $\pm$0.3} & 72.8{\scriptsize $\pm$1.8} & 86.5{\scriptsize $\pm$0.9} & 63.6{\scriptsize $\pm$1.0} & 73.2{\scriptsize $\pm$1.9} & 85.7{\scriptsize $\pm$0.9} & 71.8{\scriptsize $\pm$1.5} \\
& $p_{ _{DA}}\!=\!0.45$, $p_{ _{TR}}\!=\!0.15$ & 71.9{\scriptsize $\pm$1.1} & 86.1{\scriptsize $\pm$0.3} & \textbf{67.2{\scriptsize $\pm$0.8}} & 72.3{\scriptsize $\pm$0.9} & 86.3{\scriptsize $\pm$0.6} & \textbf{64.0{\scriptsize $\pm$1.8}} & 73.6{\scriptsize $\pm$1.5} & 86.0{\scriptsize $\pm$0.6} & \textbf{72.1{\scriptsize $\pm$0.7}} \\
\bottomrule[\heavyrulewidth]
\end{tabular}
}
\caption{\textbf{Effect of Data Augmentation Rates} (Word/Character-Level). We use different data augmentation rates ($p_{ _{DA}}$ and $p_{ _{TR}}$) and evaluate BOS\&EOS (+UNI +AUG) developed on WSJ Train/Dev. We report the macro/weighted average F1 of the BIO labeling task and the F1 score of the exact SU span extraction task.
}
\label{table:effect_augmentation}
\end{table*}

\begin{figure*}[t]
\begin{subfigure}{.33\textwidth}
  \centering
  \includegraphics[width=.99\linewidth]{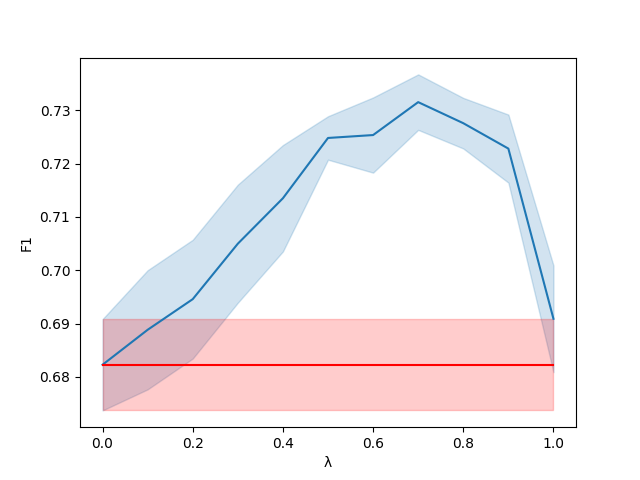}
  \caption{EWT Test ($p_{ _{CC}}\!=\!0.5$), BIO Macro}
  \label{fig:pcc05_bio_macro}
\end{subfigure}%
\begin{subfigure}{.33\textwidth}
  \centering
  \includegraphics[width=.99\linewidth]{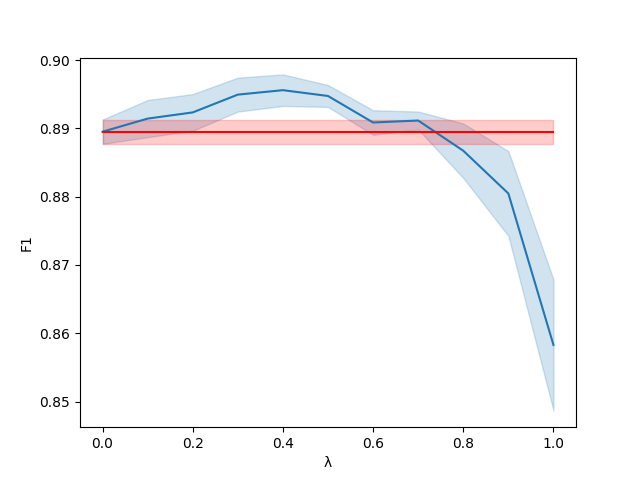}
  \caption{EWT Test ($p_{ _{CC}}\!=\!0.5$), BIO Weighted}
  \label{fig:pcc05_bio_weighted}
\end{subfigure}%
\begin{subfigure}{.33\textwidth}
  \centering
  \includegraphics[width=.99\linewidth]{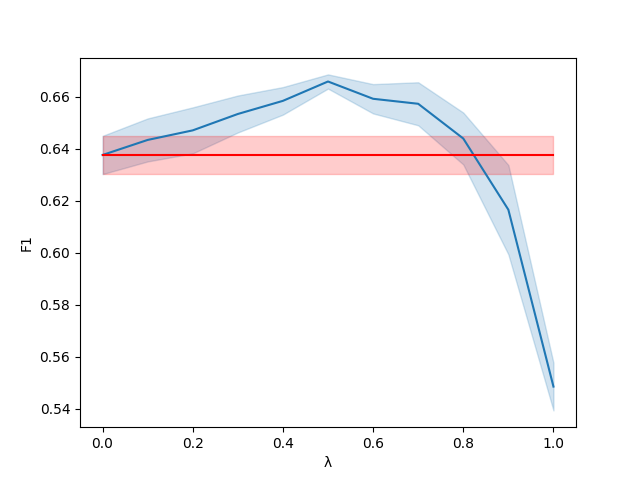}
  \caption{EWT Test ($p_{ _{CC}}\!=\!0.5$), Span}
  \label{fig:pcc05_span}
\end{subfigure}\\
\begin{subfigure}{.33\textwidth}
  \centering
  \includegraphics[width=.99\linewidth]{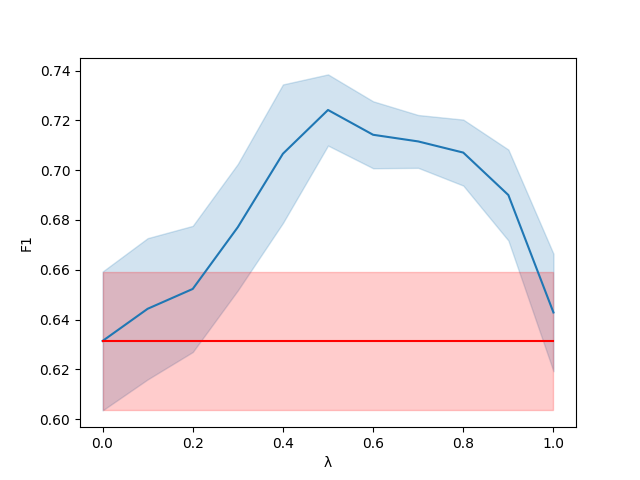}
  \caption{EWT Test ($p_{ _{CC}}\!=\!0.0$), BIO Macro}
  \label{fig:pcc00_bio_macro}
\end{subfigure}%
\begin{subfigure}{.33\textwidth}
  \centering
  \includegraphics[width=.99\linewidth]{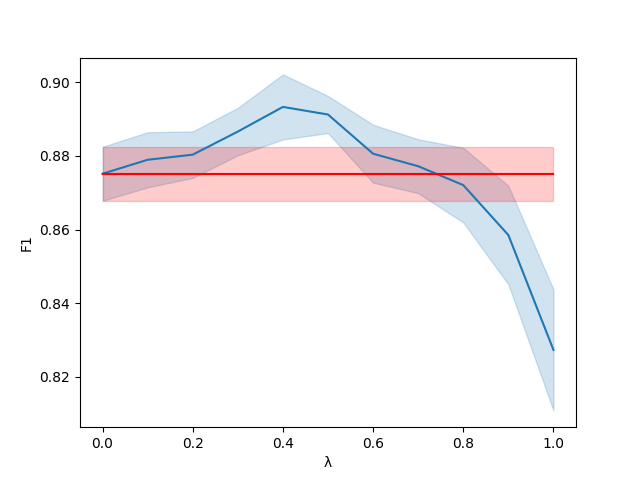}
  \caption{EWT Test ($p_{ _{CC}}\!=\!0.0$), BIO Weighted}
  \label{fig:pcc00_bio_weighted}
\end{subfigure}%
\begin{subfigure}{.33\textwidth}
  \centering
  \includegraphics[width=.99\linewidth]{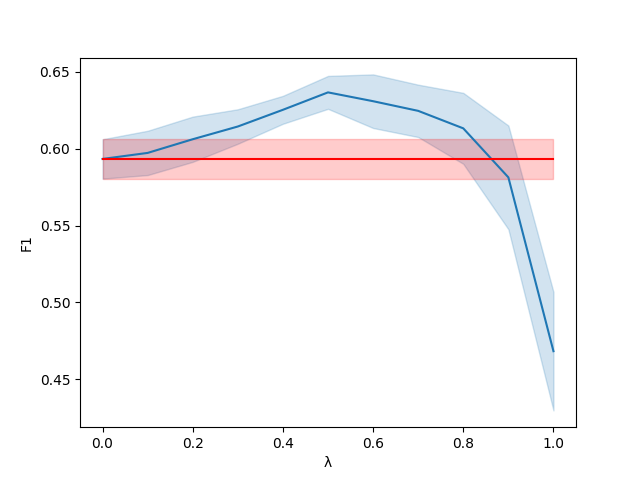}
  \caption{EWT Test ($p_{ _{CC}}\!=\!0.0$), Span}
  \label{fig:pcc00_span}
\end{subfigure}\\
\begin{subfigure}{.33\textwidth}
  \centering
  \includegraphics[width=.99\linewidth]{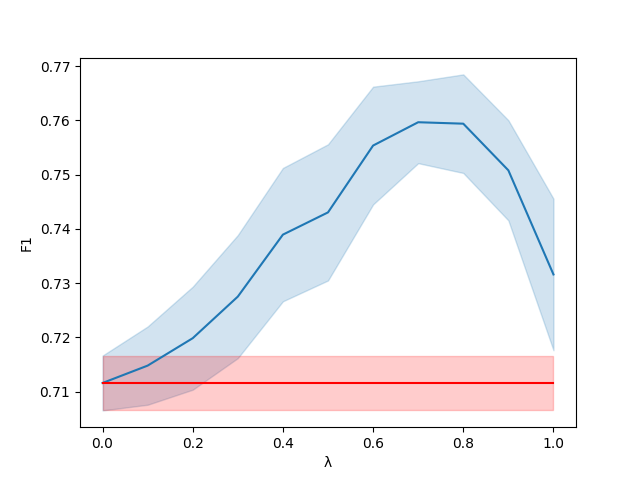}
  \caption{EWT Test (Postproc.), BIO Macro}
  \label{fig:postproc_bio_macro}
\end{subfigure}%
\begin{subfigure}{.33\textwidth}
  \centering
  \includegraphics[width=.99\linewidth]{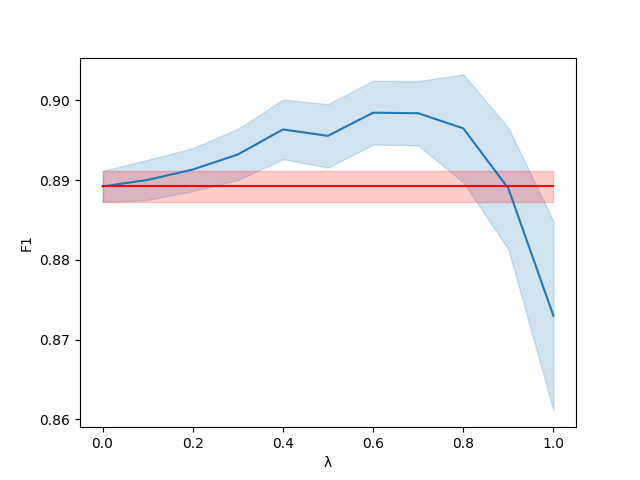}
  \caption{EWT Test (Postproc.), BIO Weighted}
  \label{fig:postproc_bio_weighted}
\end{subfigure}%
\begin{subfigure}{.33\textwidth}
  \centering
  \includegraphics[width=.99\linewidth]{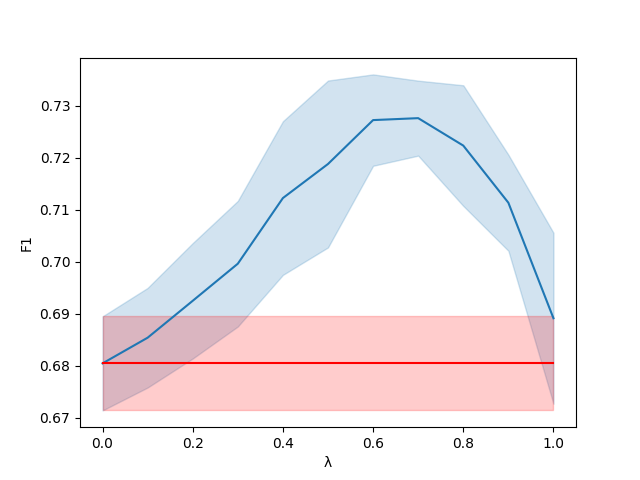}
  \caption{EWT Test (Postproc.), Span}
  \label{fig:postproc_span}
\end{subfigure}\\
\caption{\textbf{Effect of the Unidirectional Model Interpolation Rate} (Word-Level). We change $\lambda \in [0,1]$ and report the macro/weighted average F1 of the BIO labeling task and the F1 score of the exact SU span extraction task. Interpolated results are shown in blue and non-interpolated results (i.e. $\lambda = 0$) shown in red.
The line shows the mean and the shade shows the standard deviation from the five experimental runs.
}
\label{fig:effect_unidirectional}
\end{figure*}

As a default configuration, we used $p_{ _{DA}}\!=\!0.3, p_{ _{TR}}\!=\!0.1$ for the data augmentation (+AUG) and $\lambda = 0.5$ for the unidirectional model ensembling (+UNI).
To examine the effect of the choice of these hyperparameters, we conducted further experiments by changing these default hyperparameters.
Note that all evaluation results in this subsection are based on BOS\&EOS (+UNI +AUG) developed on WSJ Train/Dev.

Firstly, we focus on the data augmentation and report the results of our method trained with different sets of $p_{ _{DA}}$ and $p_{ _{TR}}$ (with $\lambda$ fixed at $0.5$).
Since increasing $p_{ _{DA}}$ leads to higher recall (and lower precision) of SU extraction and increasing $p_{ _{TR}}$ leads to higher precision (and lower recall), we used a fixed ratio of $p_{ _{DA}}:p_{ _{TR}}=3:1$ which seemed to make a good trade-off.
As shown in Table \ref{table:effect_augmentation}, the results are generally stable with the different choices of the hyperparameters.
However, more data augmentation (with larger values of $p_{ _{DA}}$ and $p_{ _{TR}}$) tends to slightly improve the performance, especially for the exact SU span extraction.

Secondly, we focus on the unidirectional model ensembling and report the results of changing the linear interpolation rate $\lambda \in [0,1]$, where $\lambda=0$ is equivalent to using only the bidirectional models and $\lambda=1$ only the unidirectional models. We fix $p_{ _{DA}}\!=\!0.3$ and $p_{ _{TR}}\!=\!0.1$ and only change $\lambda$ at the inference time without retraining the unidirectional or bidirectional models. As shown in Figure \ref{fig:effect_unidirectional}, we found that unidirectional and bidirectional models generally have complementary benefits, and choosing the intermediate value of $\lambda$ leads to the best performance. The results also indicate that we may be able to obtain further improvement by tuning $\lambda$ on the validation set, although we simply fixed $\lambda=0.5$ throughout our experiments.

\subsection{Qualitative Analyses}
\label{subsec:qualitative_analyses}

\begin{table*}[t]
\centering \scalebox{0.86}{
\begin{tabular}{cl}
\toprule[\heavyrulewidth]
 & \phantom{... \texttt{06/04/2001 05:54 PM }}\texttt{\textbf{B}} \\
Developed & ... \texttt{06/04/2001 05:54 PM \SU{Can you pass this along to Elizabeth to ensure Sanders}} \\
on EWT & \phantom{\texttt{is on board as well}}\texttt{\textbf{E}}  \\
& \texttt{\SU{is on board as well?}}\\
\midrule
 & \phantom{... \texttt{06/04/2001 05:54 PM }}\texttt{\textbf{B}} \\
Developed & ... \texttt{06/04/2001 05:54 PM \SU{Can you pass this along to Elizabeth to ensure Sanders}} \\
on WSJ & \phantom{\texttt{is on board as well}}\texttt{\textbf{E}}  \\
& \texttt{\SU{is on board as well?}}\\
\bottomrule[\heavyrulewidth]
\end{tabular}
}
\caption{\textbf{Example Outputs (Both Correct)}. We show the predictions made by our proposed method (BOS\&EOS) developed on EWT Train/Dev (top) or WSJ Train/Dev (bottom). We can verify that both methods identify the correct SU span while removing the non-sentential header as the NSU.}
\label{table:si_easy_example}
\end{table*}

\begin{table*}[t]
\centering \scalebox{0.86}{
\begin{tabular}{cl}
\toprule[\heavyrulewidth]
 & \texttt{\textbf{B}} \\
Developed & \texttt{\SU{with my breakfast I like bacon and sausage when I having a big breakfast like}} \\
on EWT & \phantom{\texttt{a grand slam with pancakes and the works}}\texttt{\textbf{E}} \\
& \texttt{\SU{a grand slam with pancakes and the works.}} \\
\midrule
& \phantom{\texttt{with my breakfast }}\texttt{\textbf{B}} \\
Developed & \texttt{with my breakfast \SU{I like bacon and sausage when I having a big breakfast like}} \\
on WSJ & \phantom{\texttt{a grand slam with pancakes and the works}}\texttt{\textbf{E}} \\
& \texttt{\SU{a grand slam with pancakes and the works.}} \\
\bottomrule[\heavyrulewidth]
\end{tabular}
}
\caption{\textbf{Example Output with One Incorrect Case}. We show the predictions made by our proposed method (BOS\&EOS) developed on EWT Train/Dev (top) or WSJ Train/Dev (bottom).
We can verify that the former extracts the correct SU span, while the latter incorrectly excludes the first prepositional phrase as an NSU.
}
\label{table:si_difficult_example}
\end{table*}

In Table \ref{table:si_easy_example} and \ref{table:si_difficult_example}, we show the actual predictions made by our proposed method developed on EWT Train/Dev and WSJ Train/Dev.
For the latter, we applied +UNI and +AUG with the default hyperparameters.

In the first example (Table \ref{table:si_easy_example}), we can verify that both models identify the correct SU span while removing the non-sentential header as the NSU.
This is a relatively easy example, since the start of the SU is capitalized and less ambiguous.

In the second example (Table \ref{table:si_difficult_example}), we can observe that our method using in-domain data (EWT Train/Dev) extracts the correct SU span, while our method developed on out-of-domain data (WSJ Train/Dev) incorrectly excludes a part of an SU.
This seems to be a relatively difficult example, since the start of the SU is not capitalized and more ambiguous.
It is worth noting that such SUs can be reliably extracted when we can leverage the in-domain annotation of gold SUs and NSUs.

\subsection{Evaluation on the Sentence Segmentation Dataset}
\label{subsec:evaluation_sentence_segmentation}

\begin{table*}[t]
\centering \scalebox{0.85}{
\begin{tabular}{crrrr}
\toprule[\heavyrulewidth]
 & & \multicolumn{1}{c}{Train} & \multicolumn{1}{c}{Dev} & \multicolumn{1}{c}{Test} \\
\midrule
\multicolumn{2}{c}{Total SUs} & 37,447 & 2,021 & 7,442 \\
\multicolumn{2}{c}{Total NSUs} & 0 & 0 & 0 \\
\midrule
\multirow{3}{*}{\shortstack[c]{Word-Level}} & B-Label & 37,447 & 2,021 & 7,442 \\
 & I-Label & 805,387 & 44,354 & 163,132 \\
 & O-Label & 0 & 0 & 0 \\
\midrule
\multirow{3}{*}{\shortstack[c]{Character-Level}} & B-Label & 37,447 & 2,021 & 7,442 \\
 & I-Label & 4,308,729 & 236,798 & 876,461 \\
 & O-Label & 0 & 0 & 0 \\
\bottomrule[\heavyrulewidth]
\end{tabular}
}
\caption{WSJ dataset statistics.}
\label{table:wsj_dataset_statistics}
\end{table*}

\begin{table*}[t]
\centering \scalebox{0.81}{
\setlength\tabcolsep{4.5pt}
\begin{tabular}{llcccccc}
\toprule[\heavyrulewidth]
\multirow{3}{*}{\shortstack[c]{Train/Dev\\Datasets}} & \multirow{3}{*}{Model} & \multicolumn{3}{c}{WSJ Test ($p_{ _{CC}}=0.5$)} & \multicolumn{3}{c}{WSJ Test ($p_{ _{CC}}=0$)} \\
\cmidrule(lr){3-5} \cmidrule(lr){6-8}
  & & BIO & BIO & \multirow{2}{*}{Span} & BIO & BIO & \multirow{2}{*}{Span} \\
 & & Macro & Weighted & & Macro & Weighted & \\
\midrule
\multirow{3}{*}{\shortstack[c]{EWT\\Train/Dev}} & EOS-Only & 97.4{\scriptsize $\pm$0.1} & 99.5{\scriptsize $\pm$0.0} & 87.3{\scriptsize $\pm$0.3} & \textbf{97.3{\scriptsize $\pm$0.0}} & 99.5{\scriptsize $\pm$0.0} & 87.2{\scriptsize $\pm$0.2} \\
 & EOS-Only (force last) & \textbf{97.6{\scriptsize $\pm$0.1}} & \textbf{99.9{\scriptsize $\pm$0.0}} & \textbf{87.8{\scriptsize $\pm$0.3}} & \textbf{97.3{\scriptsize $\pm$0.0}} & \textbf{99.6{\scriptsize $\pm$0.0}} & \textbf{87.3{\scriptsize $\pm$0.2}} \\
 & BOS\&EOS & 97.1{\scriptsize $\pm$0.2} & 99.4{\scriptsize $\pm$0.0} & 86.7{\scriptsize $\pm$0.5} & 97.0{\scriptsize $\pm$0.1} & 99.3{\scriptsize $\pm$0.0} & 86.5{\scriptsize $\pm$0.3} \\
\midrule
\multirow{6}{*}{\shortstack[c]{WSJ\\Train/Dev}} & EOS-Only & 98.4{\scriptsize $\pm$0.6} & \textbf{99.7{\scriptsize $\pm$0.1}} & 92.1{\scriptsize $\pm$2.9} & 98.2{\scriptsize $\pm$0.4} & \textbf{99.7{\scriptsize $\pm$0.1}} & 90.6{\scriptsize $\pm$1.8} \\
 & EOS-Only (force last) & 98.4{\scriptsize $\pm$0.6} & \textbf{99.7{\scriptsize $\pm$0.1}} & 92.1{\scriptsize $\pm$2.9} & 98.2{\scriptsize $\pm$0.4} & \textbf{99.7{\scriptsize $\pm$0.1}} & 90.6{\scriptsize $\pm$1.8} \\
 & EOS-Only (+AUG) & 98.2{\scriptsize $\pm$1.1} & 99.1{\scriptsize $\pm$1.0} & 92.6{\scriptsize $\pm$2.5} & 97.3{\scriptsize $\pm$1.9} & 99.3{\scriptsize $\pm$0.8} & 87.8{\scriptsize $\pm$6.3} \\
 & BOS\&EOS & \textbf{99.2{\scriptsize $\pm$0.2}} & \textbf{99.7{\scriptsize $\pm$0.3}} & \textbf{95.5{\scriptsize $\pm$0.5}} & \textbf{98.7{\scriptsize $\pm$0.1}} & \textbf{99.7{\scriptsize $\pm$0.2}} & \textbf{93.1{\scriptsize $\pm$0.4}} \\
 & BOS\&EOS (+UNI) & 98.5{\scriptsize $\pm$0.3} & 98.9{\scriptsize $\pm$0.5} & 92.9{\scriptsize $\pm$1.0} & 98.1{\scriptsize $\pm$0.3} & 98.8{\scriptsize $\pm$0.5} & 91.4{\scriptsize $\pm$0.8} \\
 & BOS\&EOS (+UNI +AUG) & 98.7{\scriptsize $\pm$0.2} & 99.3{\scriptsize $\pm$0.4} & 94.0{\scriptsize $\pm$0.7} & 98.2{\scriptsize $\pm$0.3} & 99.1{\scriptsize $\pm$0.3} & 91.8{\scriptsize $\pm$1.1} \\
\bottomrule[\heavyrulewidth]
\end{tabular}
}
\caption{\textbf{Overall Results on WSJ Test} (RoBERTa, Word-Level). We report the macro/weighted average F1 of the BIO labeling task and the F1 score of the exact SU span extraction task.
}
\label{table:roberta_wsj_results}
\end{table*}

Finally, we report the results of sentence identification on the standard sentence segmentation dataset (WSJ Test).

In Table \ref{table:wsj_dataset_statistics}, we summarize the WSJ dataset statistics.
Note that WSJ only contains SUs and do not contain any NSUs (O-labels). However, we can still evaluate the performance using the same metrics, i.e. the macro/weighted average F1 of the BIO labeling task and the F1 of the exact SU span extraction task.\footnote{Since the O-label does not exist, we report the macro average F1 as the average F1 scores of the B-label and I-label predictions.}

Table \ref{table:roberta_wsj_results} summarizes the word-level evaluation results. Since we are evaluating on WSJ Test, the performance is naturally better when the models are trained on WSJ Train/Dev rather than EWT Train/Dev (which is now out-of-domain).

When the models are trained on EWT, we found that the baseline (EOS-Only) forcing the last EOS performs the best.
This is natural, since this baseline better reflects the nature of the sentence segmentation dataset where all units are SUs.
However, our method (BOS\&EOS) is still comparable to this baseline and do not (or minimally) sacrifice performance on such datasets.

When the models are trained on WSJ, we found that our method without +UNI or +AUG performs the best. This is most likely because we can leverage the knowledge of BOS to predict EOS.
When we apply the data augmentation (+AUG) and unidirectional model ensembling (+UNI), we observe a slight decrease in performance compared to our vanilla method. However, the results are still comparable and even outperforms the baselines in some metrics (e.g. the exact SU span extraction task).

Overall, we can conclude that our methods do not sacrifice the performance on the the clean, edited texts of the sentence segmentation dataset.

\end{document}